# The Influence of Global Constraints on Similarity Measures for Time-Series Databases

Vladimir Kurbalija[1], Miloš Radovanović[1], Zoltan Geler[2], Mirjana Ivanović[1]

[1] Department of Mathematics and Informatics, Faculty of Sciences, University of Novi Sad
Trg D. Obradovića 4, 21000 Novi Sad, Serbia, {kurba,radacha,mira}@dmi.uns.ac.rs

[2] Faculty of Philosophy, University of Novi Sad
Dr Zorana Đinđića 2, 21000 Novi Sad, Serbia, gellerz@gmail.com

**Abstract.** A time series consists of a series of values or events obtained over repeated measurements in time. Analysis of time series represents an important tool in many application areas, such as stock-market analysis, process and quality control, observation of natural phenomena, medical diagnosis, etc. A vital component in many types of time-series analyses is the choice of an appropriate distance/similarity measure. Numerous measures have been proposed to date, with the most successful ones based on dynamic programming. Being of quadratic time complexity, however, global constraints are often employed to limit the search space in the matrix during the dynamic programming procedure, in order to speed up computation. Furthermore, it has been reported that such constrained measures can also achieve better accuracy. In this paper, we investigate four representative time-series distance/similarity measures based on dynamic programming, namely Dynamic Time Warping (DTW), Longest Common Subsequence (LCS), Edit distance with Real Penalty (ERP) and Edit Distance on Real sequence (EDR), and the effects of global constraints on them when applied via the Sakoe-Chiba band. To better understand the influence of global constraints and provide deeper insight into their advantages and limitations we explore the change of the 1-nearest neighbor graph with respect to the change of the constraint size. Also, we examine how these changes reflect on the classes of the nearest neighbors of time series, and evaluate the performance of the 1-nearest neighbor classifier with respect to different distance measures and constraints. Since we determine that constraints introduce qualitative differences in all considered measures, and that different measures are affected by constraints in various ways, we expect our results to aid researchers and practitioners in selecting and tuning appropriate time-series similarity measures for their respective tasks.

## 1. Introduction

A time series represents the simplest form of temporal data: a series of numbers that describes the change of the observed phenomenon over time. Each number in a time series describes the phenomenon at one point in time (Das and Gunopulos, 2003). Time series are suitable for representing social, economic and natural phenomena, medical observations, and results of scientific and engineering experiments. They are used for prediction, anomaly detection, clustering and classification, which increased the importance of different research areas of temporal data mining and resulted in a large amount of work introducing new methodologies (Das and Gunopulos, 2003; Ding et al., 2008; Han and Kamber, 2006).

The choice of an appropriate time-series similarity measure is a critical point when dealing with many tasks in mining temporal data. While working with traditional databases we are interested in data that exactly match the given query. However, in the case of similarity-based retrieval of time series, we are looking for sequences that most resemble a given series. As similarity-based retrieval is explicitly or implicitly used in all above-mentioned tasks of time-series analysis, it is important to carefully define the similarity measure between time series in order to reflect the underlying (dis)similarity of the specific data they represent (Das et al., 1997; Han and Kamber, 2006). There is a large number of (dis)similarity measures for time-series data proposed in the literature, e.g., Euclidean distance (ED) (Faloutsos et al., 1994), Dynamic Time Warping (DTW) (Keogh and Ratanamahatana, 2004), distance based on Longest Common Subsequence (LCS) (Vlachos et al., 2002), Edit Distance with Real Penalty (ERP) (Chen and Ng, 2004), Edit Distance on Real sequence (EDR) (Chen et al., 2005), Sequence Weighted Alignment model (Swale) (Morse and Patel, 2007).

Dynamic programming represents the basic technique of implementation for the vast majority of similarity measures, but because of its quadratic computational complexity it is often not suitable for larger real-world problems. To address this shortage, one can restrict the search area using global constraints such as the Sakoe-Chiba band (Sakoe and Chiba, 1978) and the Itakura parallelogram (Itakura, 1975). This can significantly speed up the calculation (Kurbalija et al., 2011). Apart from speeding up the computation it was also suggested that the usage of global constraints can actually improve the accuracy of classification compared to unconstrained similarity measures (Ratanamahatana and Keogh, 2005; Xi et al., 2006). The accuracy of classification is commonly used as a qualitative assessment of a similarity measure (Ratanamahatana and Keogh, 2005). Given all these positive effects of global constraints, it is important to carefully investigate their impact on various similarity measures.

Kurbalija et al. (2011) reported that global constraints can significantly reduce the computation time of Dynamic Time Warping and Longest Common Subsequence and that constrained measures are qualitatively different than their unconstrained counterparts. In this paper we will expand the study of the influence of the Sakoe-Chiba band on DTW and LCS and investigate its effect on two extensions of these measures: Edit Distance with Real Penalty (ERP) and Edit Distance on Real sequence (EDR). To better understand the influence of global constraints and to provide deeper insight into their advantages and limitations we will explore the change of the 1-nearest neighbor (1NN) graph with respect to the change of the constraint size. Also, we will examine how these changes reflect on the nearest-neighbors' classes and investigate their impact on the accuracy of the 1NN classifier. This choice of classifier was motivated by reports of it achieving among the best results compared to many other sophisticated classifiers for time-series data (Ding et al., 2008; Keogh, 2002; Radovanović et al., 2010; Tomašev and Mladenić, 2012; Xi et al., 2006). In addition, the accuracy of 1NN directly reflects the quality of the underlying similarity measure, the investigation of which is our primary goal.

We expect that our results will aid researchers and practitioners in selecting and tuning appropriate time-series similarity measures for their respective tasks since, as we will show, constraints introduce qualitative differences in all considered measures. Furthermore, the insight into the behavior of similarity measures with respect to changing constraints can be beneficial to the design of efficient indexing strategies for fast computation of (approximate) nearest neighbors. This statement is supported by the report (Ratanamahatana and Keogh, 2005) that the measures with the values of constraints around 5% have the same or almost the same classification accuracies as unconstrained measures. In addition, the difference of computation times between an unconstrained measure and a measure with such a small constraints is two and somewhere three orders of magnitude (Kurbalija et al., 2011).

All experiments presented in this paper are performed using the system FAP (Framework for Analysis and Prediction) (Kurbalija et al., 2010). The data for experiments is provided by the UCR Time Series Repository (Keogh et al., 2011), which includes the majority of all publicly available, labeled time-series data sets in the world.

The rest of the paper is organized as follows. Section 2 presents the necessary background knowledge about similarity measures and time-series classification and provides an overview of the related work in this area. Section 3 describes the relevant methods and tools: the Sakoe-Chiba band and the Itakura parallelogram, and the FAP system used for performing experiments. Extensive experiments and their results are given in Section 4. Section 5 concludes the paper and presents the directions for further work.

# 2. Related Work

## 2.1 Similarity Measures for Time Series

A time series of length $n$ is defined as a sequence composed of $n$ real numbers (Chen and Ng, 2004): $Q = (q_1,q_2,\ldots,q_n)$. Each element of the sequence represents the numerical value of the observed phenomena measured at a specific time. Let $Q = (q_1,q_2,\ldots,q_n)$ and $C = (c_1,c_2,\ldots,c_n)$ denote time series of the same length $n$. By interpreting time series as points in $n$-dimensional space, we can define the distance between them using the $L_p$ norm:

$$D(Q,C) = L_p(Q,C) = \sqrt[p]{\sum_{i=1}^{n} |q_i - c_i|^p}.$$

The advantage of such a similarity measure is that it represents a distance metric and can be used for indexing time series in databases. A distance metric $d: X^2 \longrightarrow \mathbf{R}$ on set $X$ is required to satisfy the following conditions:

1. $d(x,y) \geq 0$,
2. $d(x,y) = 0$ if and only if $x = y$,
3. $d(x,y) = d(y,x)$,
4. $d(x,z) \leq d(x,y) + d(y,z)$.

Due to its simplicity and speed of computation, Euclidean distance ($L_2$) has become one of the most commonly used measures of similarity between time series (Agrawal et al., 1993; Chan and Fu, 1999; Keogh et al., 2001a, 2001b). The distance between two time series is calculated based on the sum of distances between corresponding points of the series. One of the shortcomings of Euclidean distance is sensitivity to shifting and scaling along the y-axis. This deficiency can be avoided in a simple fashion by normalizing the series (Das and Gunopulos, 2003; Goldin and Kanellakis, 1995).

Another disadvantage of Euclidean distance is that the time series must be the same length and it is also sensitive to distortions and shifting along the time axis (Keogh, 2002; Ratanamahatana and Keogh, 2005). An example can be seen in Figure 1 (A) where two sequences have a similar overall shape but they are not aligned with respect to the time axis. Euclidean distance aligns the *i*-th point of the first series with the *i*-th point of the second one and will give a pessimistic estimation of their similarity. This problem can be handled using elastic distance measures such as Dynamic Time Warping and Longest Common Subsequence.

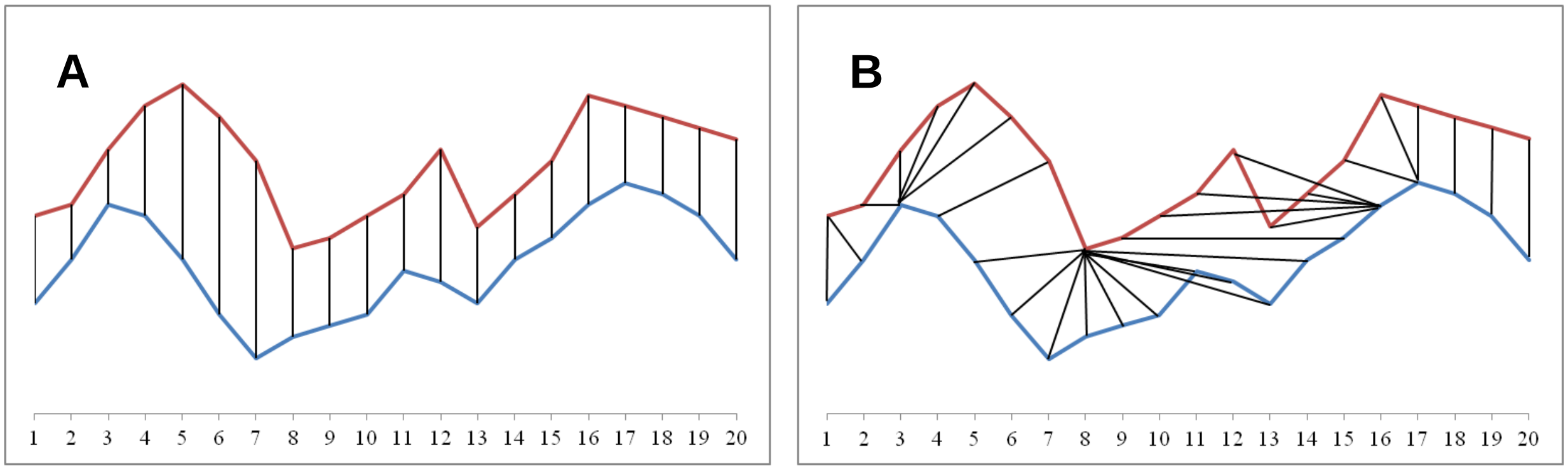

**Figure 1.** Euclidian distance and Dynamic Time Warping

Unlike Euclidean distance, Dynamic Time Warping allows non-linear aligning of the points of time series (Berndt and Clifford, 1994; Keogh, 2002; Ratanamahatana and Keogh, 2005; Xi et al., 2006). This is illustrated by the example in Figure 1 (B). DTW computes the dissimilarity by finding the optimal warping path in the matrix of distances between points of the two series as defined by Equation 1 in Figure 2. Euclidean distance represents a special case of DTW when the value of the global constraint is equal to 0.

The Longest Common Subsequence similarity measure is a variation of edit distance – the similarity between two time series is calculated as a function of the length of the longest matching subsequence (Vlachos et al., 2002). The recursive definition of LCS is given by Equation 2 in Figure 2, however the condition $q_i = c_j$ is usually too strong for time series, so it is often replaced with a parameterized condition $|q_i - c_j| \le \varepsilon$, where $0 < \varepsilon < 1$.

$$D(i,j) = \begin{cases} 0 & i = j = 0, \\ \infty & i = 0, j > 0 \text{ or } i > 0, j = 0, \\ d(q_i, c_j) + min\begin{cases} D(i-1, j-1) \\ D(i-1, j) \\ D(i, j-1) \end{cases} & i,j \ge 1. \end{cases} \quad \text{(1) DTW}$$

$$L(i,j) = \begin{cases} 0 & i = 0 \text{ or } j = 0, \\ 1 + L(i-1, j-1) & i,j > 0 \text{ and } q_i = c_j, \\ max\big(L(i-1,j), L(i,j-1)\big) & i,j > 0 \text{ and } q_i \ne c_j. \end{cases} \quad \text{(2) LCS}$$

$$E(i,j) = \begin{cases} \sum_{k=1}^{j} |c_k - g| & i = 0, \\ \sum_{k=1}^{i} |q_k - g| & j = 0, \\ min\begin{cases} |q_i - c_j| + E(i-1, j-1) \\ |q_i - g| + E(i-1, j) \\ |c_j - g| + E(i, j-1) \end{cases} & otherwise \end{cases} \quad \text{(3) ERP}$$

$$E(i,j) = \begin{cases} j & i = 0, \\ i & j = 0, \\ min\begin{cases} subcost(i,j) + E(i-1, j-1) \\ 1 + E(i-1, j) \\ 1 + E(i, j-1) \end{cases} & otherwise \end{cases} \quad \text{(4) EDR}$$

$$subcost(i,j) = \begin{cases} 0 & |q_i - c_j| \le \varepsilon, \\ 1 & |q_i - c_j| > \varepsilon. \end{cases}$$

**Figure 2.** Definitions of elastic distance measures

DTW solves the problem of local time shifting of time series (Chen and Ng, 2004; Das and Gunopulos, 2003; Keogh, 2002), can work with time series of different lengths (Berndt and Clifford, 1994), but since it does not satisfy the triangle inequality it is not a distance metric (Yi et al., 1998). Furthermore, like Euclidean distance, it is relatively sensitive to noise (Vlachos et al., 2002). The similarity measure Edit distance with Real Penalty is based on $L_1$ and DTW measures. It solves the problem of local time shifting and represents a distance metric (Chen and Ng, 2004). ERP is defined by Equation 3 in Figure 2. This similarity measure introduces a constant value $g$ (with the default value $g = 0$) as the gap of the edit distance and uses $L_1$ distance between elements as the penalty to handle local time shifting (Chen and Ng, 2004; Chen et al., 2005). The shortage of ERP is that it is also sensitive to noise (Chen et al., 2005). ERP was successfully used in solving various problems including classification of pulse waveforms (Zhang et al., 2010), clustering trajectories of moving objects (Pelekis et al., 2009), and querying time-series streams (Gopalkrishnan, 2008).

Taking into account only sufficiently similar points, LCS solves the problem of the presence of noise (Vlachos et al., 2002), but does not differentiate similar time series consisting of similar sub-series with different sizes of gaps between them (Chen et al., 2005) and does not satisfy the triangle inequality (Vlachos et al., 2002). The similarity measure Edit Distance on Real sequence is based on edit distance like LCS, but unlike LCS, it finds the minimal number of edit operations needed to convert one time series to another (Chen et al., 2005). This measure is defined by Equation 4 in Figure 2. EDR is robust to noise and is expected to be more accurate than LCS because it assigns penalties to gaps between two similar sub-series according to the lengths of the gaps, but is not a distance metric (Chen et al., 2005). A novel clustering method of trajectories based on EDR is presented in (Abul et al., 2010).

## 2.2 Time-Series Classification

Time-series classification has attracted much attention recently in the time-series community. Approaches to classification vary from purely statistical methods such as exponential smoothing or ARIMA models, to those based on different data-mining techniques like neural networks, genetic algorithms, support vector machines and fuzzy systems.

Statistical methods usually use Autoregressive (AR) models where the current value of time series is generated as a linear combination of the previous values. These methods are more appropriate for forecasting, but some interesting works can be found in the area of classification. Kini and Sekhar (2013) present the large margin autoregressive (LMAR) method that uses an AR model for each class and the large margin method for estimation of parameters of AR models. A system which builds groups of time series that share the same forecasting model applied to supply chains is presented by Turrado García et al. (2012). The similarity between two time series is defined in the following manner: two series will have the same associated ARIMA model if and only if the autocorrelation and partial autocorrelation functions give similar results in their $N$ first positions. In the paper by Huan and Palaniappan (2004) the neural-network classification of autoregressive features is applied on time series of electroencephalogram signals extracted during mental tasks.

An interesting approach for time-series classification is to transform the time series into standard feature vectors with a fixed dimensionality. In this case, many well-studied data-mining techniques can then be adopted for classification or clustering. Zhang et al. (2006) proposed an algorithm based on the orthogonal wavelet transform, in which the coefficients of the Haar wavelet were extracted as a feature vector for subsequent time-series clustering. Eruhimov et al. (2007) use DTW to transform the time axis of each signal in order to decrease the Euclidean distance between signals from the same class. Afterwards, a range of attributes of both transformed time series and original time series are extracted to form a high-dimensional feature vector. Another interesting approach is the adaptation of segment-based representations to extract features from time series (Geurts and Wehenkel, 2005; Lee et al., 2008; Megalooikonomou et al., 2005). Recent activities in transformation of time series into feature vectors include utilization of segment-base features (Zhang et al., 2012) and the extraction of meaningful patterns from original data (Zhang et al., 2009).

The most widely used approach to time-series classification is to define the distance function between two time series and use some of the existing distance-based classifiers. In this approach, the key problem is how to define a robust distance or similarity measure that can reflect the overall shape of the time series. The most standard distance measures are described in the previous subsection. Recently, an alignment-based distance metric called Time Warp Edit Distance (TWED) was proposed by Marteau (2009). It has been proved that this metric satisfies the triangle inequality. Górecki and Luczak (2013) emphasize the importance of using derivatives in time-series distance functions. This approach considers the overall shape of a time series rather than individual point-to-point function comparison. A generalization of the DTW measure is proposed by Jeong et al. (2011) as a novel distance measure, called Weighted DTW (WDTW). This measure penalizes the points according to the phase difference between a reference point of the first time series and test point of the second time series. The proposed approach can prevent some bad alignments where one point of the first time series maps onto a large part of the second time series.

In this paper we adopt the aforementioned widely used approach to time-series classification of using distance measures between time series in conjunction with an existing distance-based classifier – 1NN. As was mentioned in the introduction, this choice of classifier was motivated by reports of it achieving among the best results (Ding et al., 2008; Keogh, 2002; Radovanović et al., 2010; Tomašev and Mladenić, 2012; Xi et al., 2006). The additional upside of 1NN is that its accuracy directly reflects the quality of the underlying distance measure, which augments our qualitative analysis of the impact of constraints, and also provides a practical demonstration of our observations.

# 3. Methods and Tools

## 3.1 Global Constraints of Time-Series Distance Measures

Each of the above-mentioned elastic similarity measures (DTW, LCS, ERP and EDR) relies on dynamic programming for finding the optimal path within the search matrix. Dynamic programming requires comparing each element of one time series with each element of the other one. This makes the calculation of the similarity measures quite slow and has some limitations when dealing with large data sets. To improve the performance of these algorithms a number of techniques have been developed. The Sakoe-Chiba band (Sakoe and Chiba, 1978) narrows the warping window around the diagonal of the matrix using a constant range *r*. The Itakura parallelogram (Itakura, 1975) uses a similar approach: the range of the restriction is a function of *i* and *j* coordinates in the matrix. These restrictions of the search path are illustrated in Figure 3. Global constraints were originally introduced to prevent some bad alignments, where a relatively small part of one time series maps onto a large section of another time series.

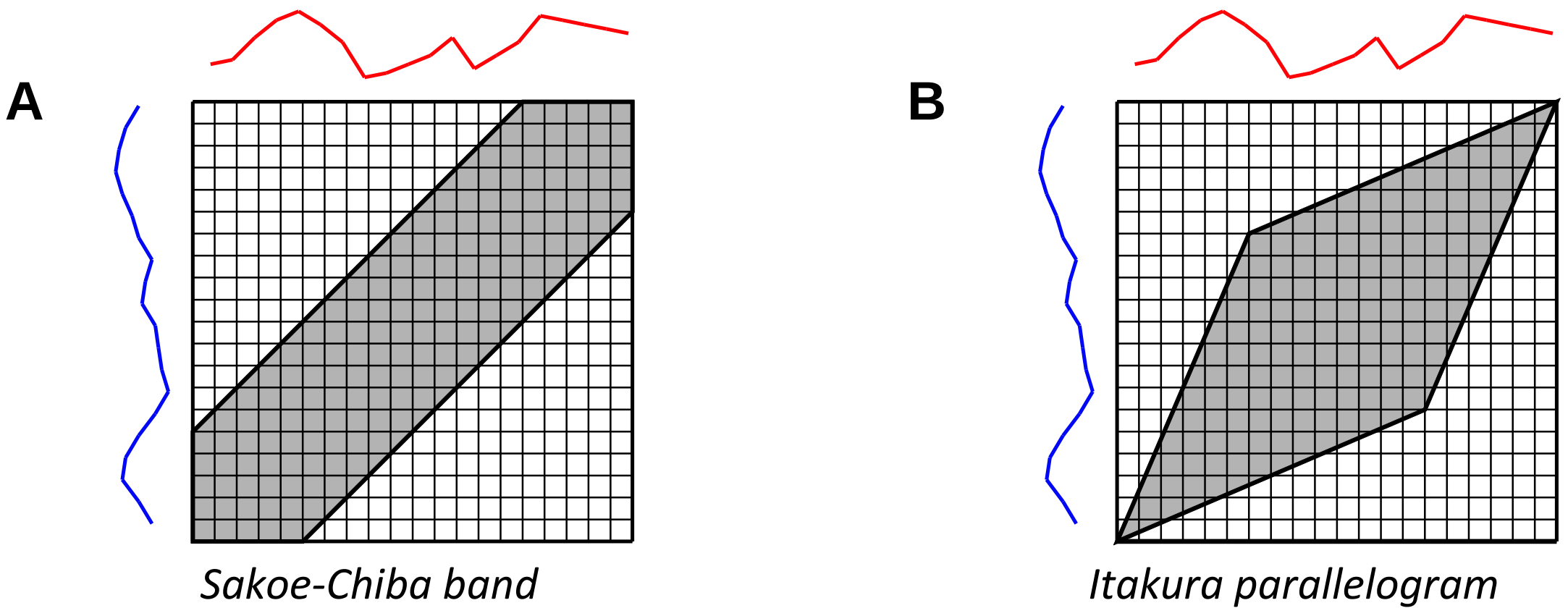

**Figure 3.** Sakoe-Chiba band and Itakura parallelogram

Apart from speeding up the computation it was also suggested that the use of global constraints can actually improve the accuracy of classification compared to unconstrained similarity measures (Ratanamahatana and Keogh, 2005; Xi et al., 2006). Recently, Kurbalija et al. (2011) showed that global constraints can significantly reduce the computation time of DTW and LCS and that constrained measures are qualitatively different from their unconstrained counterparts. In Section 4, we will explore a wide variety of *r* values and examine their effect on DTW, LCS, ERP and EDR distance measures constrained by the Sakoe-Chiba band.

## 3.2 The FAP System

The usefulness of time series in the study of social, economic and natural phenomena has resulted in the development of a number of tools which, through different approaches, enable the analysis time series in various ways. Existing software systems for time-series analysis can be classified into two main groups: general data-mining and machine-learning systems (like WEKA (Hall et al., 2009) and RapidMiner (Rapid-I, 2013)), and statistical systems supporting statistical and econometric models of time series (like the SAS software (SAS, 2013) and GRETL (Baiocchi and Distaso, 2003)). Besides these two large groups of applications, there are also specialized programs for summarization and visualization of time series such as Spiral (Weber et al., 2001) and VizTree (Lin et al., 2004).

The increased importance of studying different research areas of temporal data mining has resulted in a large amount of work introducing new methodologies for different task types including indexing, classification, clustering, prediction, segmentation, anomaly detection, etc. (Das and Gunopulos, 2003; Ding et al., 2008; Han and Kamber, 2006; Vogrinčič and Bosnić, 2011). When dealing with time series there are several important concepts which need to be considered: pre-processing transformation, time-series representation, and similarity measures. The task of the pre-processing is to remove different kinds of distortions from raw time series. Reducing high dimensionality of time series by preserving its important properties is the task of different time-series representations: Discrete Fourier Transformation (DFT), Singular Value Decomposition (SVD), Discrete Wavelet Transformation (DWT), Piecewise Aggregate Approximation (PAA), Adaptive Piecewise Constant Approximation (APCA), etc. The task of the (dis)similarity measure between time series is to reflect the underlying (dis)similarity of the specific data they represent. There is a number of distance measures for similarity of time-series data: $L_p$ distance, Dynamic Time Warping, distance based on Longest Common Subsequence, Edit Distance with Real Penalty, Edit Distance on Real sequence, Sequence Weighted Alignment model (Swale), etc.

All these concepts, when introduced, are usually separately implemented and presented in different publications. Every newly-introduced representation method or distance measure has claimed a particular superiority (Ding et al., 2008). However, this was usually based on comparison with only a few counterparts of the proposed concept. On the other hand, to the best of our knowledge there is no freely available system for time-series analysis and mining which supports all mentioned concepts, with the exception of the work proposed in (Ding et al., 2008).

The motivation behind developing the Framework for Analysis and Prediction (FAP) system (Kurbalija et al., 2010) as a multipurpose and multifunctional library which supports the above mentioned important concepts of time-series analysis is to integrate these techniques into one common framework. The FAP library is designed to be a free, extensible open-source software package that implements the main techniques and methods needed for the analysis of time series (pre-processing, similarity measures, representation) and for temporal data mining (indexing, classification, prediction, etc.). By developing the FAP library in the form of an open source, free, and extensible framework we wanted to facilitate and accelerate the exploration of various new techniques of time-series analysis and mining, and to help researchers in comparing newly introduced and proposed concepts with the existing ones. The framework can be also useful to non-professionals from different domains as an assistance tool for choosing appropriate methods for their own data sets.

In the current state of development, all main similarity measures are implemented, as well as several classifiers, statistical tests and representations. The implemented similarity measures include $L_p$, Swale, unconstrained and constrained DTW, LCS, ERP and EDR. The constrained measures are implemented using the Sakoe-Chiba band and the Itakura parallelogram. The system contains the implementation of the 1NN and kNN classifiers, the Holdout, Cross-Validation and Leave-One-Out testing methods and the following time-series representations: PLA, PAA, APCA, SAX and Spline (Kurbalija et al., 2009).

## 4. The Impact of the Sakoe-Chiba Band on Time-Series Distance Measures

In our previous work (Kurbalija et al., 2011), we have studied the impact of the Sakoe-Chiba band on the speed of computing Dynamic Time Warping and Longest Common Subsequence, observing 38 data sets. We have also examined the change of the nearest neighbor graph using a small number of different warping window widths (75%, 50%, 25%, 20%, 15%, 10%, 5%, 1% and 0% of the length of time series). Based on these preliminary results we concluded that in case of DTW and LCS the constrained measures, besides better performance, represent qualitatively different measures than the unconstrained ones.

In this section we will report the results of our expanded study of the influence of the Sakoe-Chiba band on the most widely used elastic similarity measures: DTW, LCS, ERP and EDR. To better understand the influence of global constraints we will explore the efficiency and behavior of the 1NN classifier for different values of constraints and investigate its accuracy. We will report the change of the 1NN graph with regard to the change of the global constraints. Our choice of 1NN was mainly motivated by reports that it achieves among the best results compared to many other sophisticated classifiers for time-series data (Ding et al., 2008; Keogh, 2002; Radovanović et al., 2010; Tomašev and Mladenić, 2012; Xi et al., 2006). In addition, the accuracy of 1NN directly reflects the quality of the underlying similarity measure.

In the first phase of the experiments we will explore the change of the 1-nearest neighbor graph with respect to the change of the constraint size (Section 4.1). In the second phase we will investigate how these changes impact on the 1NN classifier regarding the nearest-neighbors' classes (Section 4.2). The examination of the constraint's impact on classification accuracy is discussed in the third part of our study (Section 4.3). For all four considered similarity measures we have limited the warping window with the following constraint values for *r*: 90%, 80%, 70%, 60%, 50%, 45%, 40%, 35%, 30%, and all values from 25% to 0% in steps of 1% of the time-series length. These values were chosen because it is expected that the measures with larger constraints behave similarly to the unconstrained measure, while smaller constraints have more interesting behavior (Kurbalija et al., 2011; Ratanamahatana and Keogh, 2005; Xi et al., 2006). Furthermore, we included eight new data sets in the experiments.

The experiments were conducted on 46 data sets from (Keogh et al., 2011), which includes the majority of all publicly available, labeled time-series data sets in the world. The properties of these data sets are shown in Table 1. The length of time series varies from 24 to 1882 depending of the data set. The number of time series per data set varies from 56 to 9236 and the number of classes varies from 2 to 50. The data sets originate from a plethora of different domains, including medicine, robotics, astronomy, biology, face recognition, handwriting recognition, etc.

| Data set | Size | Length | Classes | Data set | Size | Length | Classes | Data set | Size | Length | Classes |
|---|---|---|---|---|---|---|---|---|---|---|---|
| 50words | 905 | 270 | 50 | fish | 350 | 463 | 7 | sonyaiborobotsurface | 621 | 70 | 2 |
| adiac | 781 | 176 | 37 | gun_point | 200 | 150 | 2 | sonyaiborobotsurfaceii | 980 | 65 | 2 |
| beef | 60 | 470 | 5 | haptics | 463 | 1092 | 5 | starlightcurves | 9236 | 1024 | 3 |
| car | 120 | 577 | 4 | inlineskate | 650 | 1882 | 7 | swedishleaf | 1125 | 128 | 15 |
| cbf | 930 | 128 | 3 | italypowerdemand | 1096 | 24 | 2 | symbols | 1020 | 398 | 6 |
| chlorineconcentration | 4307 | 166 | 3 | lighting2 | 121 | 637 | 2 | synthetic_control | 600 | 60 | 6 |
| cinc_ecg_torso | 1420 | 1639 | 4 | lighting7 | 143 | 319 | 7 | trace | 200 | 275 | 4 |
| coffee | 56 | 286 | 2 | mallat | 2400 | 1024 | 8 | twoleadecg | 1162 | 82 | 2 |
| cricket_x | 780 | 300 | 12 | medicalimages | 1141 | 99 | 10 | twopatterns | 5000 | 128 | 4 |
| cricket_y | 780 | 300 | 12 | motes | 1272 | 84 | 2 | uwavegesturelibrary_x | 4478 | 315 | 8 |
| cricket_z | 780 | 300 | 12 | noninvasivefatalecg_thorax1 | 3765 | 750 | 42 | uwavegesturelibrary_y | 4478 | 315 | 8 |
| diatomsizereduction | 322 | 345 | 4 | noninvasivefatalecg_thorax2 | 3765 | 750 | 42 | uwavegesturelibrary_z | 4478 | 315 | 8 |
| ecg200 | 200 | 96 | 2 | oliveoil | 60 | 570 | 4 | wafer | 7164 | 152 | 2 |
| ecgfivedays | 884 | 136 | 2 | osuleaf | 442 | 427 | 6 | wordssynonyms | 905 | 270 | 25 |
| faceall | 2250 | 131 | 14 | plane | 210 | 144 | 7 | yoga | 3300 | 426 | 2 |
| facefour | 112 | 350 | 4 | | | | | | | | |

**Table 1.** Properties of the data sets

### 4.1. Change of the 1NN Graph with Narrowing Constraints

The nearest-neighbor graph is a directed graph where each time series is connected with its nearest neighbor. We calculated this graph for unconstrained measures and for measures with the following constraints: 90%, 80%, 70%, 60%, 50%, 45%, 40%, 35%, 30%, and all values from 25% to 0% in steps of 1% of the time-series length. After that, we observed the change of the nearest neighbor graphs as the percentage of time series (nodes in the graph) that changed their nearest neighbor compared to the nearest neighbor in the unconstrained measure. The graphical representation of results can be seen in Figures 4 through 7 for DTW, LCS, ERP and EDR, respectively. Each figure is represented by two charts for the sake of readability. The first chart (A) contains the behavior of 10 most representative data sets, illustrating the behavior of the majority of data sets. The second chart (B) shows the general statistics over all data sets: minimum values, maximum values, average values and the deviations from the average values.

The 1NN graphs of the DTW measure (Figure 4) remain the same until the size of the constraint is narrowed to approximately 60%–50%, and after that the graphs start to change. As the width of the warping window becomes smaller, an increasing number of data sets exhibits bigger changes. When the size of the Sakoe-Chiba band falls below 5% of the time-series length, changes are present in all of the data sets. For r = 0%, changes higher than 50% have been registered for the majority of the data sets (the only exceptions are *beef*, *chlorineconcentration*, *coffee*, *italypowerdemand*, *mallat* and *oliveoil*) and for some of them the alteration levels even reach values above 90% (*50words*, *cbf*, *starlightcurves*, *synthetic_control*, *twopatterns*, *uwavegesturelibrary_x*, *wordssynonyms*).

The situation with LCS (Figure 5) is even more drastic: the 1NN graphs remain the same to approximately 30%–25%, while for smaller constraints they change more quickly for most of the data sets. When *r* reaches 0%, changes greater than 90% occur in a much larger number of data sets than in the case of DTW (17, opposed to 7). However, there are some exceptions. A number of data sets (*beef*, *ecgfivedays*, *mallat*, *oliveoil, trace*) exhibits changes only for very small values of the constraint (less than 2%) and in the case of *chlorineconcentration* there are some oscillations.

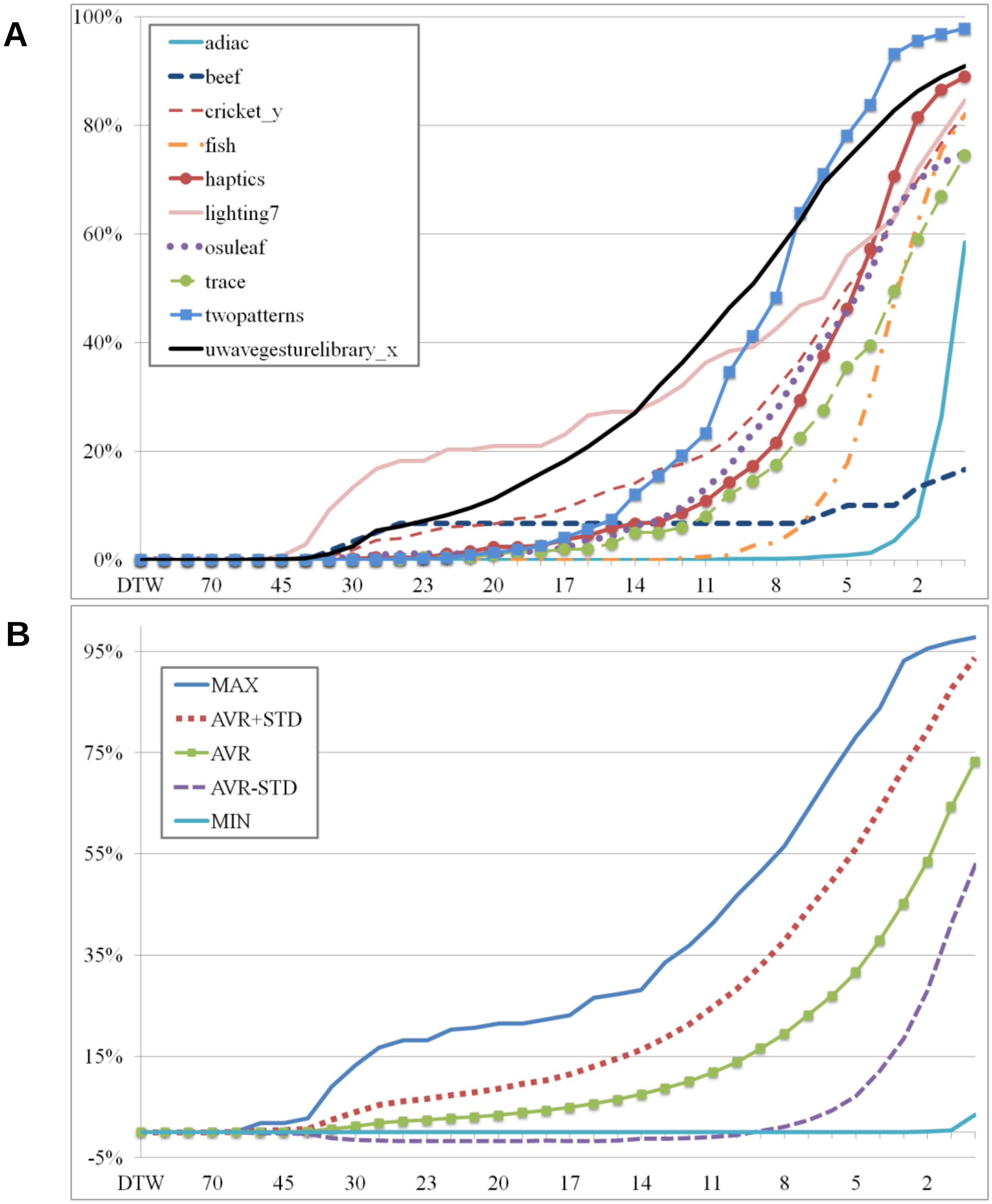

**Figure 4.** Change of 1NN graph for DTW

A

B

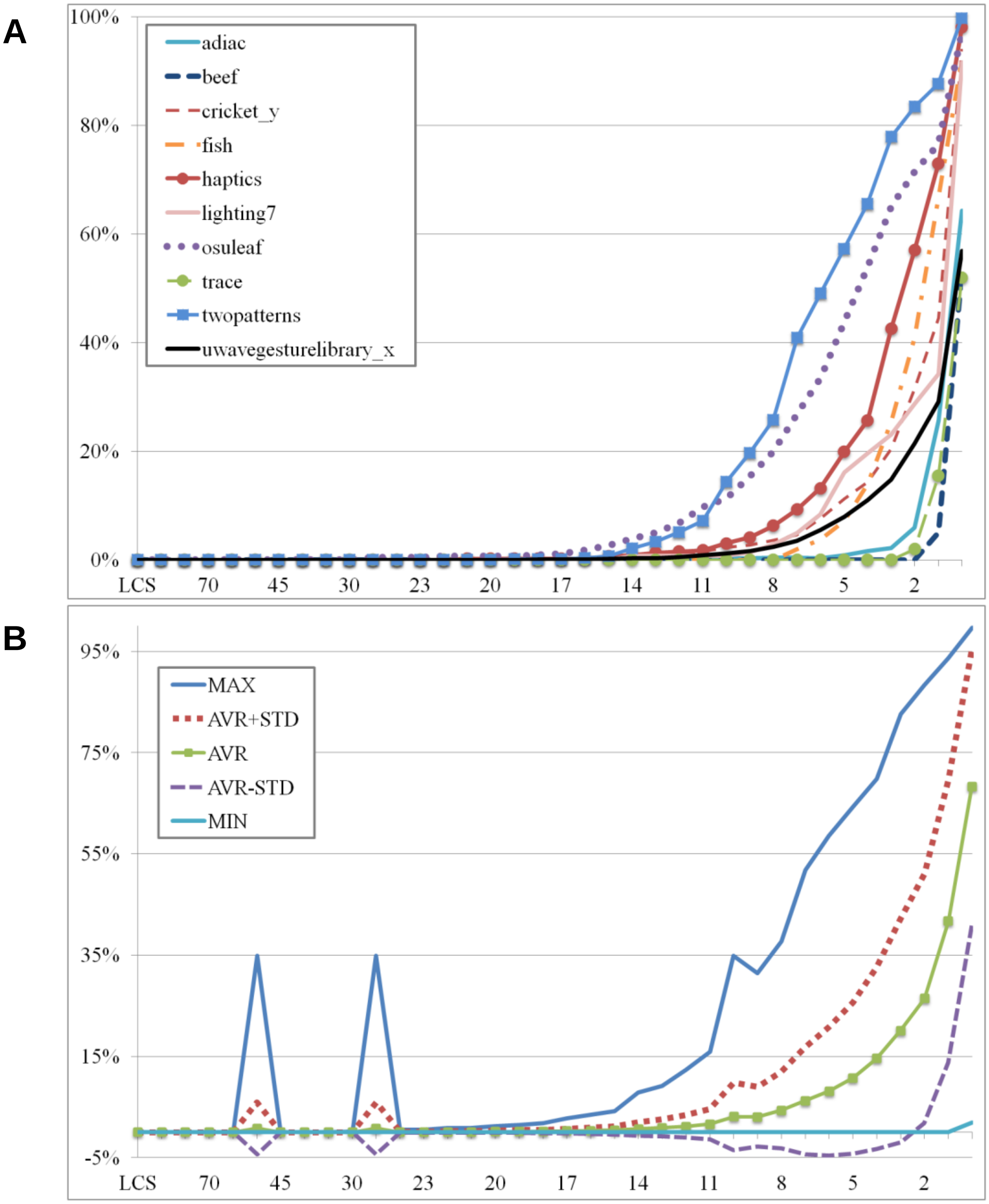

**Figure 5.** Change of 1NN graph for LCS

The 1NN graph of the ERP measure (Figure 6) remains the same until the size of the constraint is narrowed to approximately 60%–50%, similarly as in the case of DTW. After that the graph starts to change more noticeably. For small values of the constraint (5%–0%) this change becomes significant for most of the data sets and in some cases even reaches values above 70%–90%. It is also evident that the use of the Sakoe-Chiba band does not affect the 1NN graph in the same way for each data set: in case of a small number of data sets the changes are subtle or there are no changes at all (*adiac*, *chlorineconcentration*, *coffee*, gun_*point*, *mallat*, *oliveoil*, *trace*).

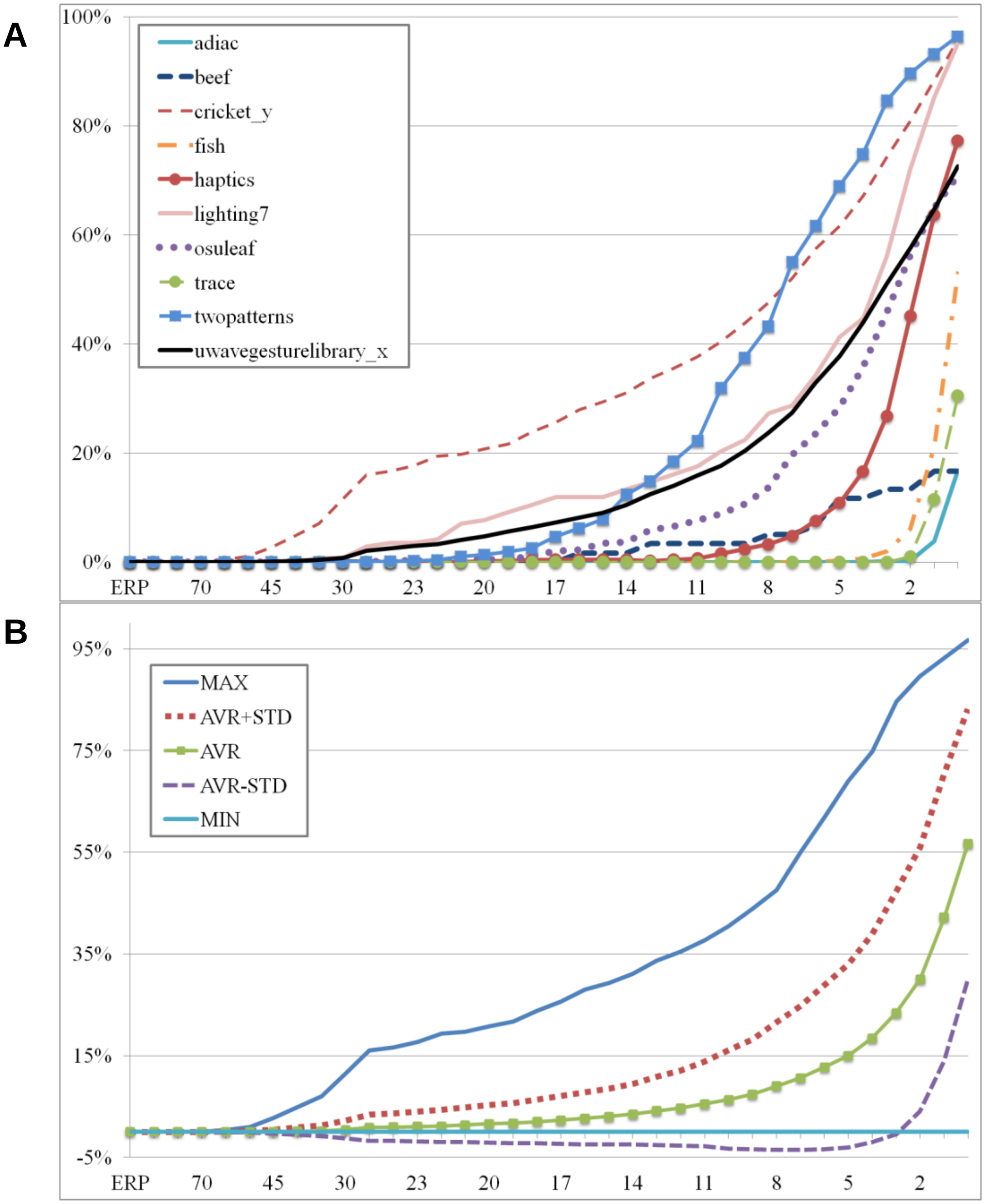

**Figure 6.** Change of 1NN graph for ERP

EDR (Figure 7) behaves in a similar manner to LCS, but there are three noticeable differences. The changes begin later, only when the value of the constraint drops below 20%. The changes do not reach such high values as in the case of LCS – the maximum values are between 60%–80%. Again, there is a small number of data sets where the changes are subtle (*beef*, *chlorineconcentration*, *ecgfivedays*, *mallat*, *oliveoil*, *trace*).

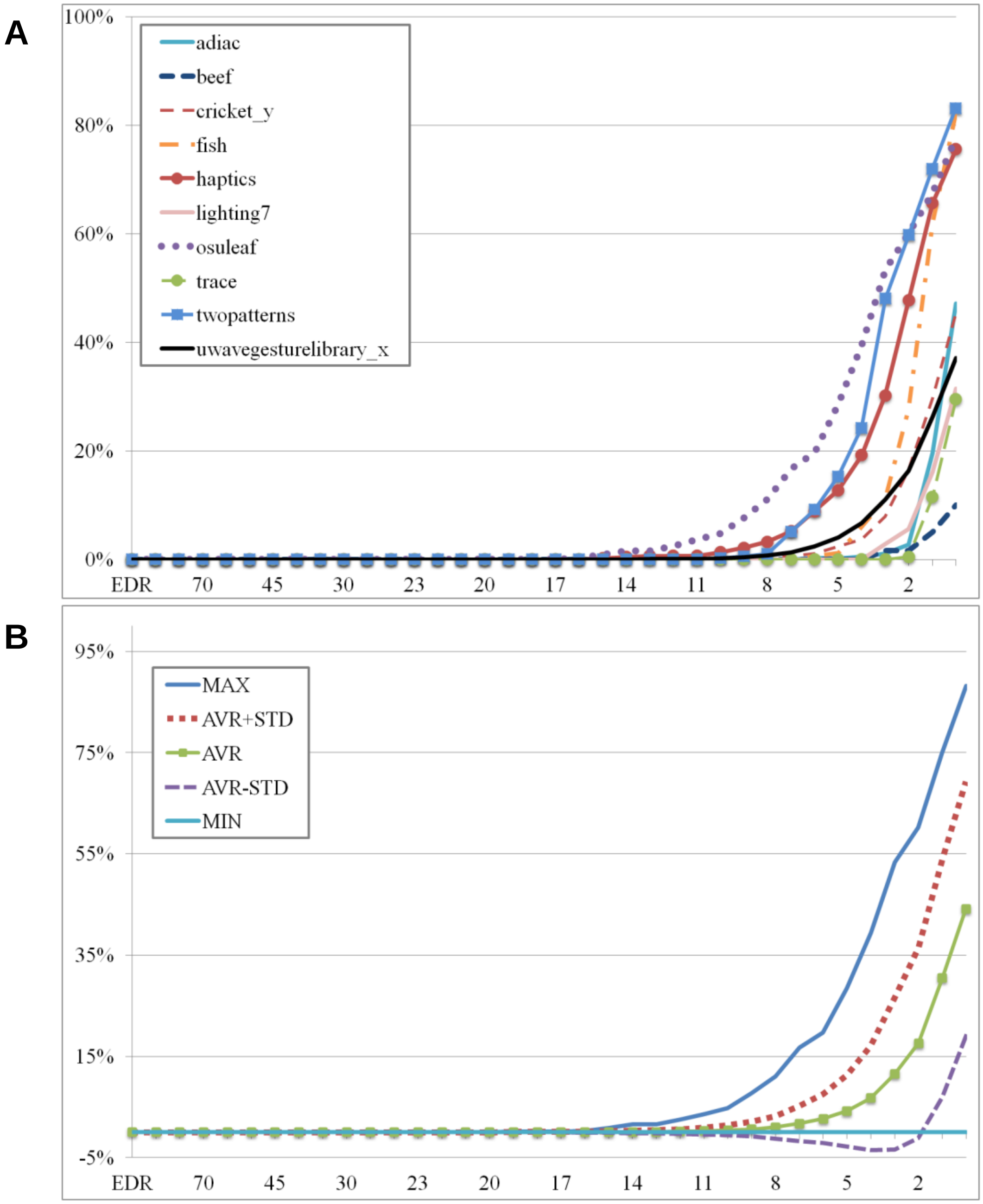

**Figure 7.** Change of 1NN graph for EDR

From the obtained results we can clearly see that for low values of the constraint the change of the 1NN graph becomes significant for most data sets in the case of all four similarity measures.

This observation suggests that the constrained measures represent qualitatively different measures than the unconstrained ones. It is also clear that the application of the Sakoe-Chiba band does not have the same effect on all data sets, and there are noticeable differences in the behavior of the similarity measures. By studying the results we can immediately see two quite conspicuous characteristics of the Sakoe-Chiba band. First, warping window width of 0% most drastically affects LCS: for a significant number of data sets there is a sudden increase in the percentage of changed neighbors compared to $r$ = 1% (some examples are: *cinc_ecg_torso*, *coffee*, *diatomsizereduction*, *noninvasivefatalecg_thorax1*, *noninvasivefatalecg_thorax2* and *wafer*). Secondly, LCS, ERP and EDR for some data sets (for example, *chlorineconcentration*, *mallat*, *oliveoil*, *trace*) show only tenuous changes (or no changes at all except for the values of constraint $r$ = 0%), with appreciably higher changes in the case of DTW.

The general traits of differences between the similarity measures can be easily seen on the charts in Figure 8 and Figure 9. Figure 8 presents the average values of the changes in the 1NN graphs across all data sets, and Figure 9 shows the percentages of those data sets for which there are changes in the 1-nearest neighbor graph produced by the constrained similarity measures, compared to the 1NN graph of the unconstrained ones. It is obvious that the use of the Sakoe-Chiba band exhibits the greatest influence on DTW: changes in the 1NN graph arise as soon as the size of the constraint is narrowed to 60% and for very narrow warping windows they reach (on average) the highest values among the observed similarity measures. The smallest influence occurs in the case of EDR: the 1NN graph begins to change only when the size of the warping window drops below 20% of the length of time series, and the average change for $r$ = 0% is lowest here. For most data sets LCS and ERP behave very similarly: they are situated "between" DTW and EDR. This relationship between the similarity measures can also be seen in Figure 10 which shows the highest width of the warping window required to change at least 10% of the nodes in the 1NN graph (the first chart again contains 10 most representative data sets for the sake of readability, while the second chart shows general statistics for all data sets). Changes of this magnitude appear earliest for DTW (with average warping window width about 10.18%), followed by ERP (6.33%), then by LCS (4.78%), and at the end by EDR (2.54%). Comparisons using the Wilcoxon sign-rank test (García et al., 2009) reveal statistical significance of pairwise differences, with $p$-values shown in Table 2 (where the difference between ERP and LCS may be considered the one borderline case).

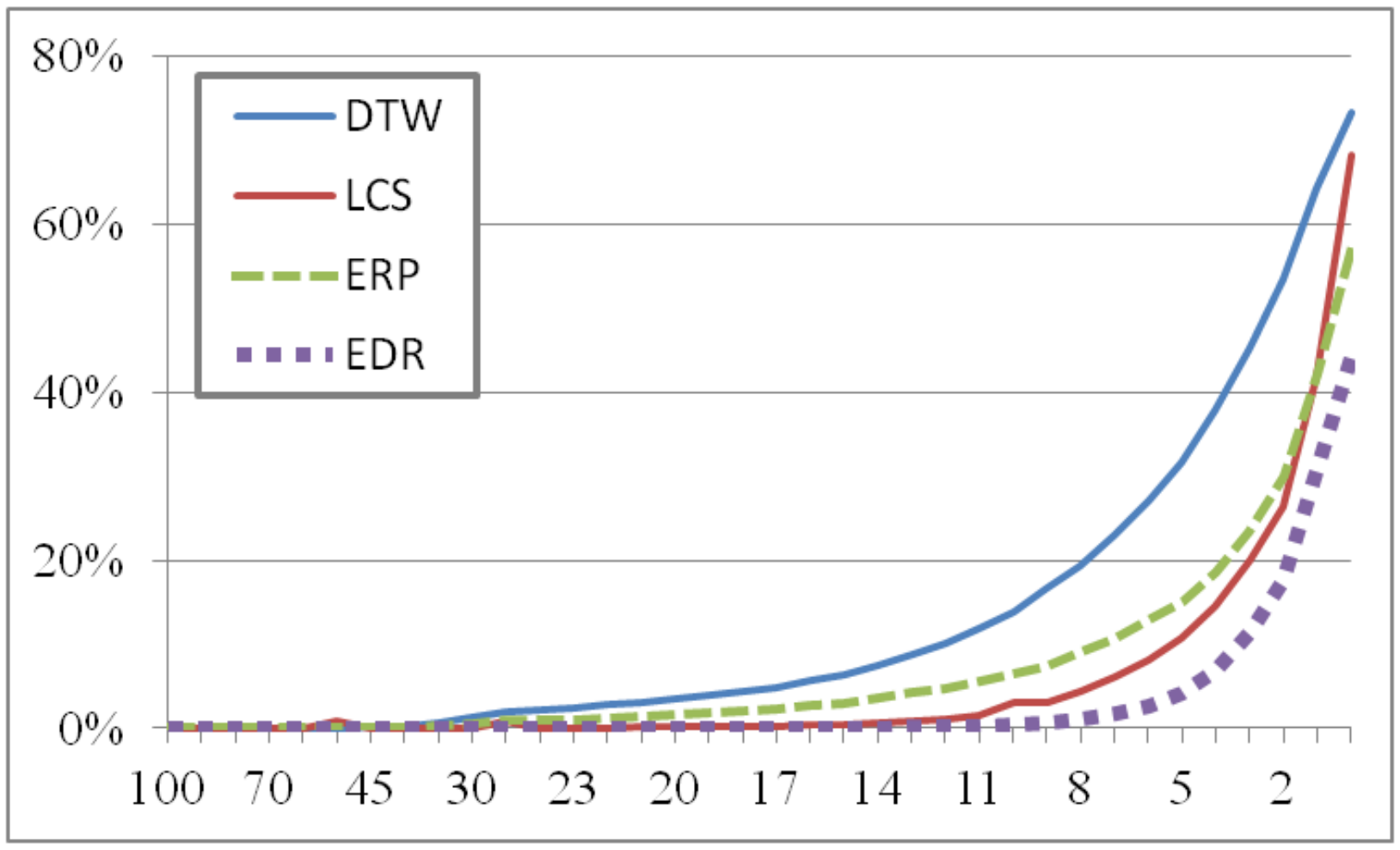

**Figure 8.** The average changes in the 1NN graph

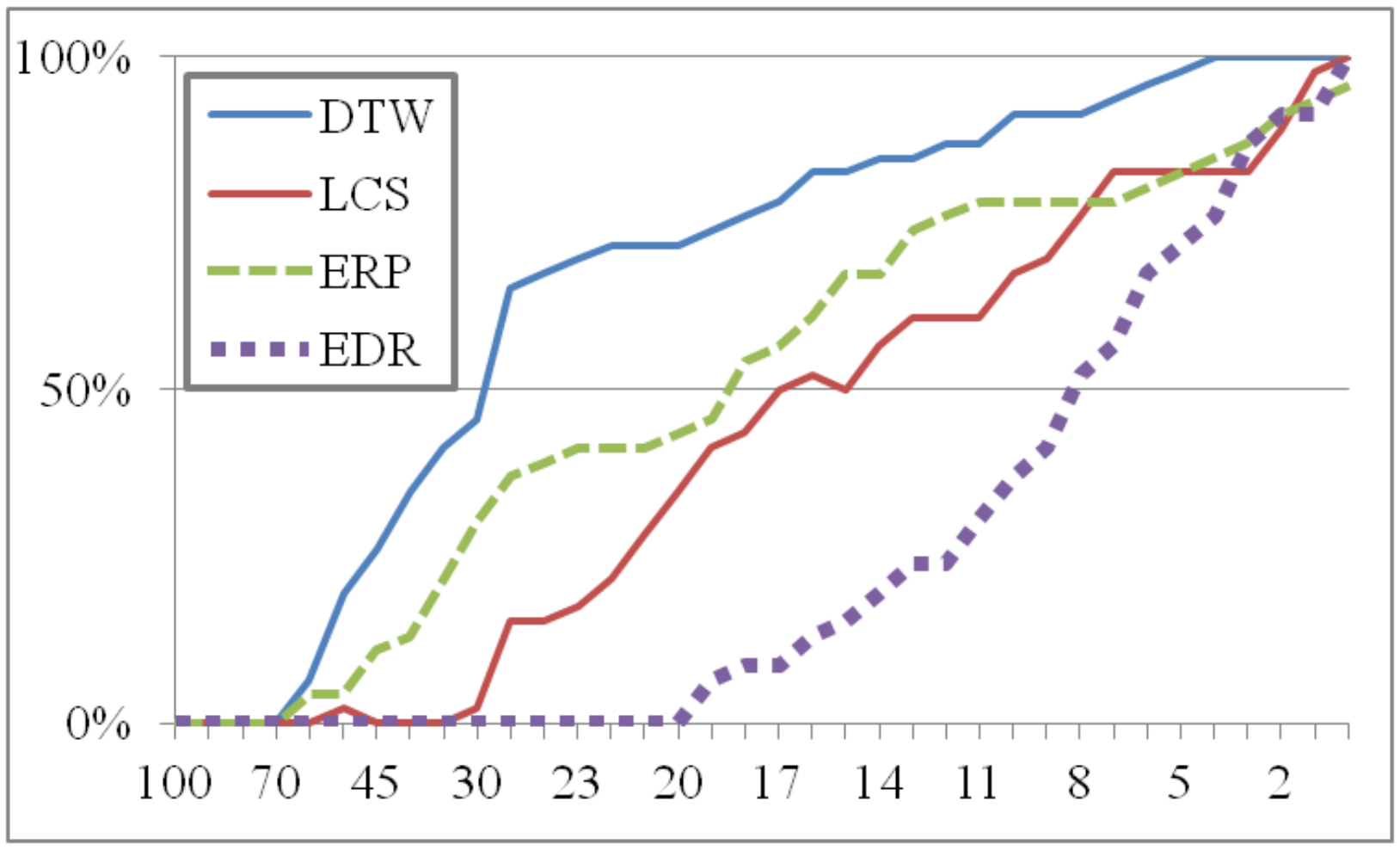

**Figure 9.** The percentage of the data sets with changed 1NN graphs

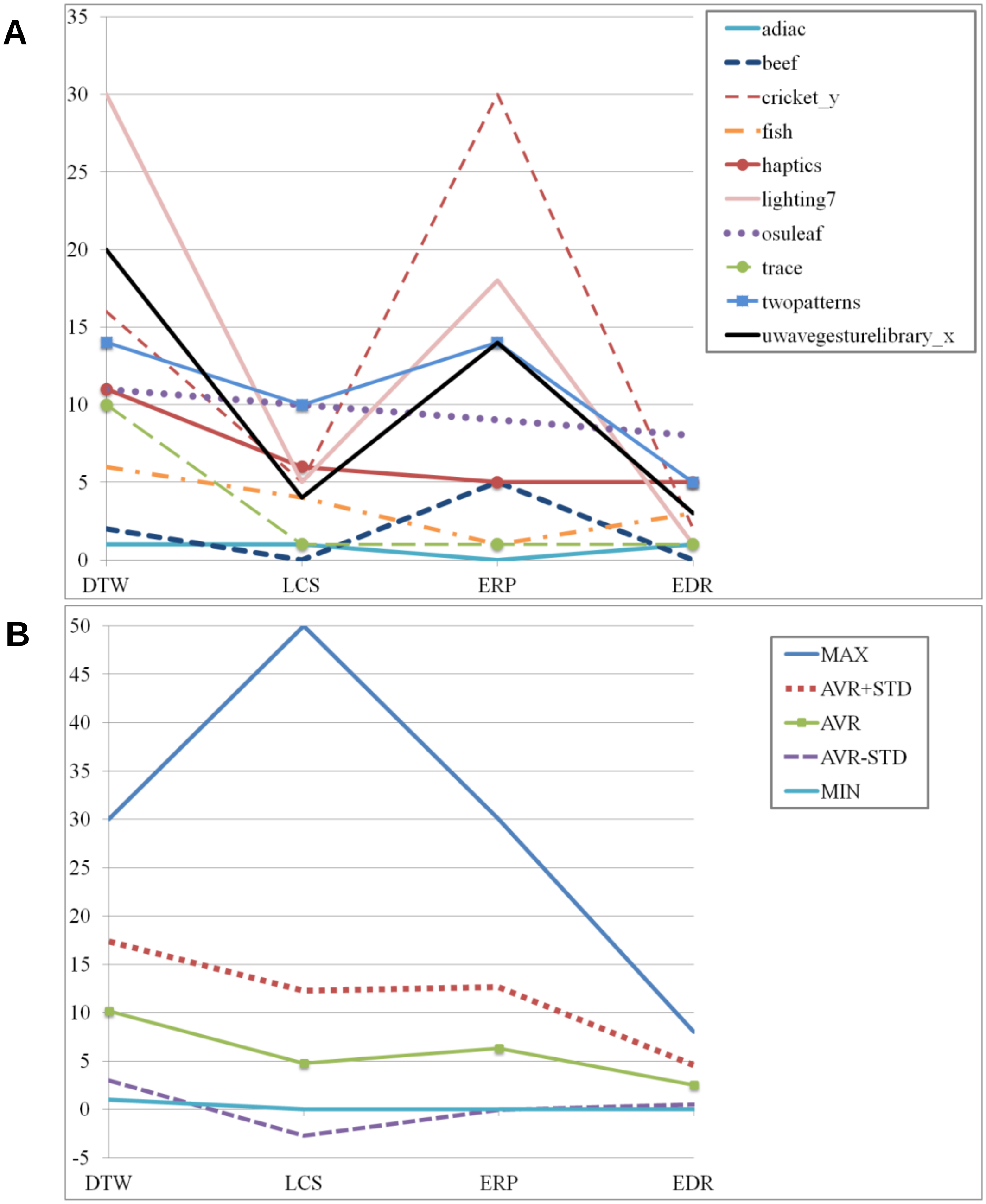

**Figure 10.** The smallest warping windows needed to change at least 10% of the 1NN graph

| | LCS | ERP | EDR |
|---|---|---|---|
| DTW | 4.63E-07 | 3.63E-06 | 9.25E-09 |
| LCS | | 0.018192 | 0.001279 |
| ERP | | | 2.16E-05 |

**Table 2.** *p* values for the pairwise Wilcoxon sign-rank test of the differences in the highest width of the warping window required to change at least 10% of the nodes in the 1NN graph

## 4.2 Change of Classes with Narrowing Constraints

The results described in Section 4.1 clearly indicate that the application of the Sakoe-Chiba band can significantly change the structure of the 1-nearest neighbor graph, especially for small warping-window widths. We have also seen that this influence is not manifested in the same way for different similarity measures. In this section we give an overview of how these changes in the 1-nearest neighbor graph may affect the behavior of the 1NN classifier.

Classification denotes the process of grouping time series into predefined classes. The 1NN classifier represents a very simple form of classification: the class of the unclassified time series is determined as the class of its most similar time series. Despite its simplicity, the 1NN classifier often produces better results than other more complex classifiers (Ding et al., 2008; Keogh, 2002; Radovanović et al., 2010; Xi et al., 2006).

Since the results of 1NN classification depend entirely on the class of the nearest neighbor, changes in the nearest-neighbor graph directly affect the classification. In this section we will examine the extent to which nearest neighbors change their classes under the influence of the Sakoe-Chiba band. Similarly to the first part of the experiments, we record the percentage of those nearest neighbors which have under the influence of global constraints changed their classes compared to the nearest neighbors in the unconstrained measure. The graphical representation of results can be seen in Figures 11 through 14 for DTW, LCS, ERP and EDR, respectively. Each figure is represented by two charts for the sake of readability. The first chart (A) contains the behavior of 10 most representative data sets while the second chart (B) shows the general statistics over all data sets: minimum values, maximum values, average values and the deviation from the average values.

In Section 4.1 we have seen that major changes to the 1NN graph occur for $r < 5\%$. In this section we will examine that particular area in more detail. For easier description of the obtained results we rely on the following notation. Let $N$ denote the set of nodes in the 1NN graph that changed their nearest neighbor compared to the nearest neighbor in the unconstrained measure, and let $C$ denote the set of nodes in the 1NN graph whose nearest neighbor changed its class compared to the class of the nearest neighbor in the unconstrained measure. Obviously, $C$ is a subset of $N$. Let $\delta$ denote the fraction of those modified nodes that have also changed their class:

$$\delta = \frac{|C|}{|N|}$$

The values of $\delta$ for $r < 5\%$ are given in Tables 3 through 6 for DTW, LCS, ERP and EDR. Within these tables, data sets with the highest $\delta$ values (greater than or equal to 50%) are marked with symbol ●, and those with the lowest values (smaller than or equal to 10%) with symbol ○. The dash sign among the results in these tables indicates that there are no changes in the nearest neighbor graph compared to the unconstrained similarity measure.

In case of DTW (Figure 11), nearest neighbors with changed classes are beginning to appear immediately with the first changes in the structure of the 1NN graph: when the width of the warping window drops to about 60% of the length of time series. The percentage of neighbors with changed class increases as the width of the Sakoe-Chiba band narrows, and for some data sets reaches values higher than 40% of the number of time series in the data set (*haptics*, *inlineskate*). Looking at Table 3 we can see that for $r < 5\%$ on average only about 22% of the changed nodes have modified their classes compared to the unconstrained measure. Changes greater than 50% are only present for three data sets: *adiac*, *haptics* and *inlineskate*. On the other hand, for one fourth of the observed data sets, $\delta$ is less than 10%. This is somewhat surprising since in the first part of the experiments we have found significant changes in the structure of the 1NN graph for $r = 0\%$ (about 25%–98%) for all data sets (except *chlorineconcentration* and *beef*).

| | Data set | 4% | 3% | 2% | 1% | 0% | | Data set | 4% | 3% | 2% | 1% | 0% |
|---|---|---|---|---|---|---|---|---|---|---|---|---|---|
| | 50words | 39.57 | 38.75 | 38.88 | 41.11 | 43.49 | ○ | *mallat* | *3.64* | *3.66* | *3.68* | *3.34* | *3.09* |
| ● | adiac | **60.00** | **74.07** | **58.06** | **54.85** | **50.44** | | medicalimages | 31.71 | 31.91 | 34.72 | 34.35 | 34.35 |
| | beef | 16.67 | 16.67 | 25.00 | 33.33 | 50.00 | | motes | 15.52 | 15.51 | 13.87 | 13.77 | 13.77 |
| | car | 50.00 | 42.00 | 27.54 | 28.21 | 34.12 | | noninvasivefatalecg_thorax1 | 37.11 | 31.88 | 29.34 | 27.16 | 25.26 |
| ○ | *cbf* | *0.00* | *0.13* | *0.00* | *0.11* | *1.02* | | noninvasivefatalecg_thorax2 | 28.73 | 24.56 | 23.47 | 20.58 | 15.87 |
| ○ | *chlorineconcentration* | *0.00* | *0.00* | *0.00* | *0.00* | *0.00* | | oliveoil | 0.00 | 33.33 | 28.57 | 25.00 | 11.76 |
| ○ | *cinc_ecg_torso* | *9.85* | *7.58* | *4.68* | *2.66* | *1.62* | | osuleaf | 50.00 | 47.54 | 46.28 | 47.06 | 50.45 |
| | coffee | 25.00 | 42.86 | 41.67 | 46.15 | 42.86 | | plane | 33.33 | 20.00 | 10.00 | 7.41 | 4.17 |
| | cricket_x | 32.45 | 31.62 | 30.80 | 31.43 | 41.69 | ○ | *sonyaiborobotsurface* | *5.70* | *5.70* | *4.05* | *2.61* | *2.61* |
| | cricket_y | 31.02 | 31.02 | 31.93 | 34.89 | 42.66 | | sonyaiborobotsurfaceii | 11.06 | 8.01 | 8.01 | 5.96 | 5.96 |
| | cricket_z | 31.88 | 30.82 | 32.76 | 32.59 | 42.60 | | starlightcurves | 9.22 | 9.19 | 10.13 | 11.10 | 12.49 |
| | diatomsizereduction | 33.33 | 25.00 | 8.33 | 1.52 | 0.40 | | swedishleaf | 30.24 | 27.52 | 28.65 | 29.00 | 32.02 |
| | ecg200 | 25.00 | 21.84 | 21.70 | 18.52 | 18.52 | ○ | *symbols* | *1.45* | *1.50* | *1.59* | *2.56* | *3.33* |
| ○ | *ecgfivedays* | *2.04* | *1.27* | *0.90* | *1.28* | *1.36* | ○ | *synthetic_control* | *2.07* | *2.09* | *2.09* | *8.24* | *8.24* |
| | faceall | 13.53 | 6.81 | 5.12 | 4.88 | 5.90 | | trace | 0.00 | 0.00 | 2.54 | 5.22 | 14.77 |
| | facefour | 27.27 | 16.67 | 12.90 | 10.00 | 8.33 | ○ | *twoleadecg* | *0.00* | *0.29* | *0.35* | *0.45* | *0.45* |
| | fish | 41.67 | 36.75 | 29.95 | 28.03 | 29.62 | ○ | *twopatterns* | *0.00* | *0.04* | *0.10* | *0.35* | *1.06* |
| | gun_point | 22.54 | 17.65 | 16.00 | 12.95 | 12.00 | | uwavegesturelibrary_x | 27.15 | 27.08 | 27.14 | 27.23 | 27.59 |
| ● | haptics | **69.06** | **65.75** | **67.11** | **67.33** | **68.93** | | uwavegesturelibrary_y | 37.83 | 37.21 | 36.86 | 37.06 | 37.38 |
| ● | inlineskate | **62.73** | **64.21** | **64.53** | **66.23** | **64.23** | | uwavegesturelibrary_z | 32.77 | 32.57 | 32.55 | 33.63 | 34.45 |
| ○ | *italypowerdemand* | *5.45* | *5.45* | *5.45* | *5.45* | *5.45* | ○ | *wafer* | *3.10* | *2.57* | *2.23* | *1.13* | *0.70* |
| | lighting2 | 12.16 | 17.95 | 22.58 | 23.23 | 29.13 | | wordssynonyms | 37.71 | 36.99 | 36.59 | 38.89 | 41.33 |
| | lighting7 | 32.94 | 38.89 | 36.89 | 39.29 | 43.80 | | yoga | 20.44 | 17.62 | 13.53 | 10.79 | 9.60 |

| | 4% | 3% | 2% | 1% | 0% |
|---|---|---|---|---|---|
| **Average** | 23.11% | 22.84% | 21.29% | 21.24% | 22.37% |

**Table 3.** $\delta$ values for DTW

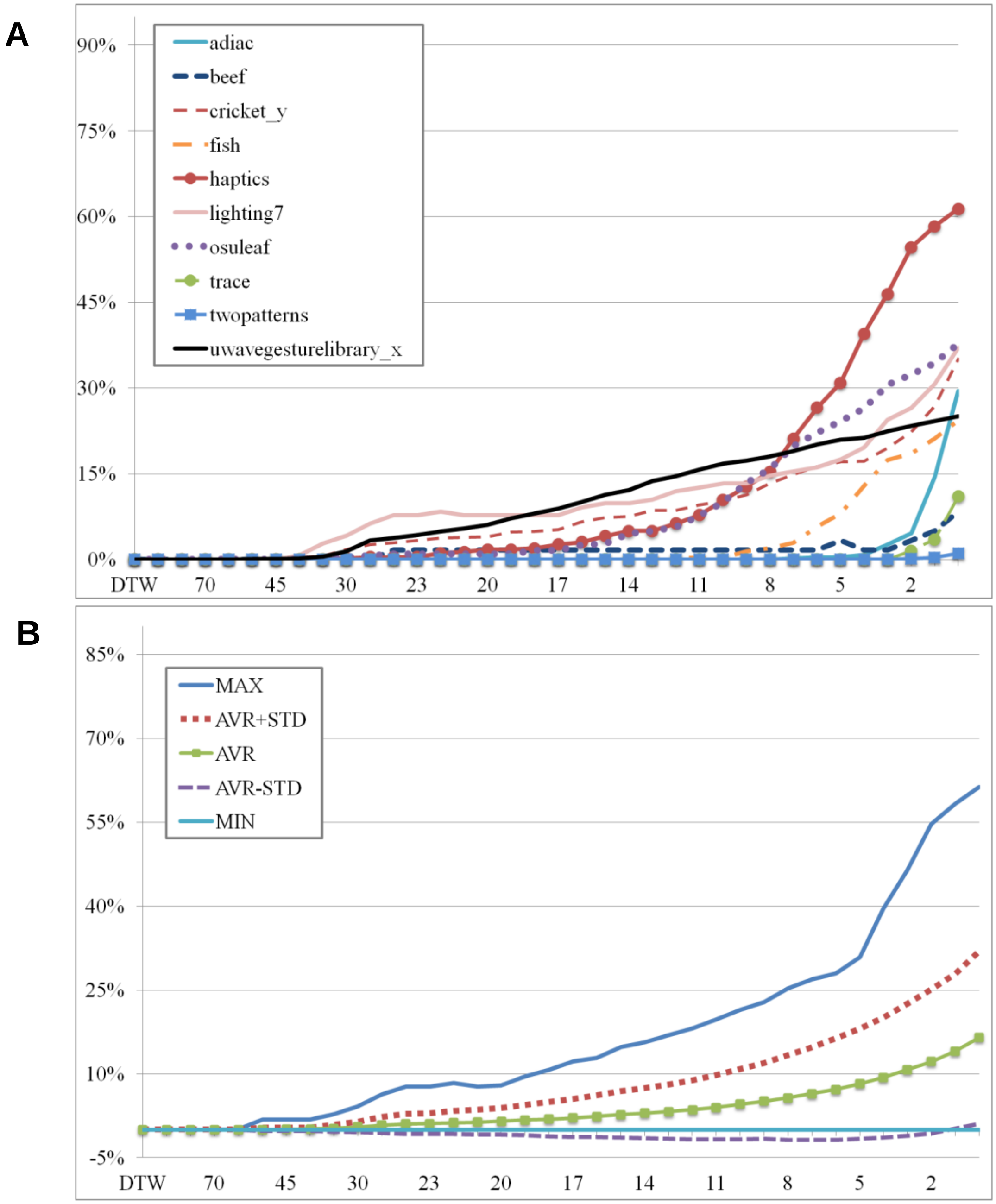

**Figure 11.** Change of classes for DTW

In accordance with the structure of the nearest neighbor graph for LCS, neighbors with altered classes begin to occur when the width of the warping window reaches 25%, but their number starts to grow significantly only when it drops below 10% (Figure 12). For *r* = 0% the changes are more noticeable than with DTW, for a number of data sets they even reach values above 70% (*cricket_x*, *cricket_y*, *cricket_z*). For *lighting2*, *starlightcurves*, *symbols*, *synthetic*_control, and *yoga* major changes in the 1NN graph at *r* = 0% (greater than 90%) are accompanied by significantly smaller changes in terms of classes (smaller than 40%). This indicates that only a smaller part of the changed nodes also changed their class. Comparing the results in Table 4 and Table 3 we can see that the average δ values for LCS are noticeably higher than for DTW, especially for $r$ = 0%, where it is almost twice as high (the Wilcoxon sign-rank test indicates significance for $r$ = 1% and $r$ = 0%, with $p$ values of 0.025467 and 3.34E-08, respectively). This means that changes in the structure of the LCS 1NN graph induced by applying the Sakoe-Chiba band more significantly alter the classes of nodes than for DTW. Another noticeable difference between these two similarity measures refers to the presence of greater fluctuations of δ values for some data sets using LCS (*cbf*, *coffee*, *diatomsizereduction*, *faceall*, *facefour*, *synthetic_control,* and *yoga*). There is also some resemblance between LCS and DTW: *adiac*, *haptics* and *inlineskate* are again among the data sets with the highest δ values, and some of the data sets with the lowest δ values are common for these two measures.

| Data set | 4% | 3% | 2% | 1% | 0% | Data set | 4% | 3% | 2% | 1% | 0% |
|---|---|---|---|---|---|---|---|---|---|---|---|
| 50words | 43.55 | 39.66 | 37.74 | 35.83 | 58.88 | mallat | - | - | - | 0.00 | 3.33 |
| ● **adiac** | **69.23** | **76.47** | **78.26** | **64.00** | **60.36** | medicalimages | 41.44 | 38.07 | 37.50 | 55.18 | 55.18 |
| ● **beef** | **-** | **-** | **-** | **66.67** | **84.38** | motes | 14.29 | 10.91 | 7.17 | 20.05 | 20.05 |
| car | 42.86 | 28.57 | 23.81 | 21.25 | 74.11 | noninvasivefatalecg_thorax1 | 87.50 | 100.00 | 72.00 | 50.50 | 39.83 |
| cbf | 0.46 | 0.52 | 1.34 | 2.53 | 62.66 | noninvasivefatalecg_thorax2 | 85.71 | 90.00 | 83.33 | 54.95 | 29.94 |
| chlorineconcentration | - | - | - | 16.78 | 9.58 | ● **oliveoil** | **-** | **-** | **-** | **100.00** | **64.29** |
| ○ *cinc_ecg_torso* | - | - | *0.00* | *2.94* | *4.81* | osuleaf | 35.71 | 36.11 | 35.13 | 39.64 | 64.72 |
| coffee | - | - | 50.00 | 100.00 | 25.00 | plane | 11.11 | 15.38 | 4.55 | 2.50 | 8.64 |
| cricket_x | 58.82 | 52.87 | 42.49 | 39.52 | 79.01 | sonyaiborobotsurface | 0.00 | 0.00 | 3.85 | 14.55 | 14.55 |
| cricket_y | 41.96 | 35.85 | 31.43 | 31.99 | 81.42 | sonyaiborobotsurfaceii | 23.53 | 27.12 | 27.12 | 17.47 | 17.47 |
| cricket_z | 55.08 | 50.00 | 45.56 | 39.83 | 79.42 | starlightcurves | 11.34 | 11.34 | 11.44 | 11.34 | 18.19 |
| diatomsizereduction | 100.00 | 66.67 | 40.00 | 7.69 | 9.43 | swedishleaf | 12.32 | 11.74 | 14.16 | 17.11 | 35.84 |
| ecg200 | 30.30 | 26.67 | 19.74 | 22.29 | 22.29 | symbols | 3.55 | 3.80 | 4.24 | 4.86 | 39.46 |
| ecgfivedays | - | - | - | - | 23.53 | synthetic_*control* | 8.47 | 7.05 | 7.05 | 31.32 | 31.32 |
| faceall | 18.31 | 5.65 | 4.28 | 4.05 | 28.50 | trace | - | - | 50.00 | 45.16 | 15.38 |
| facefour | 40.00 | 12.50 | 7.41 | 2.78 | 51.43 | ○ *twoleadecg* | *0.00* | *0.00* | *0.00* | *3.82* | *3.82* |
| fish | 46.94 | 32.58 | 28.67 | 27.35 | 63.16 | twopatterns | *0.12* | *0.13* | *0.26* | *0.78* | *70.22* |
| ○ *gun_point* | *0.00* | *0.00* | *4.55* | *3.85* | *8.40* | uwavegesturelibrary_x | 44.40 | 42.21 | 38.28 | 34.20 | 43.11 |
| ● **haptics** | **77.31** | **66.50** | **65.53** | **65.09** | **71.37** | uwavegesturelibrary_y | 42.24 | 37.26 | 35.55 | 34.16 | 53.97 |
| ● **inlineskate** | **51.39** | **59.50** | **64.13** | **63.16** | **64.94** | uwavegesturelibrary_z | 38.26 | 35.57 | 33.51 | 34.22 | 60.30 |
| ○ *italypowerdemand* | *10.00* | *10.00* | *10.00* | *10.00* | *10.00* | wafer | 50.00 | 22.22 | 12.82 | 1.34 | 1.01 |
| lighting2 | 30.00 | 36.00 | 22.58 | 22.22 | 40.00 | wordssynonyms | 40.63 | 36.93 | 34.48 | 33.27 | 61.74 |
| lighting7 | 60.71 | 60.61 | 46.34 | 44.90 | 66.41 | yoga | 19.73 | 10.11 | 9.97 | 8.74 | 31.07 |

| | 4% | 3% | 2% | 1% | 0% |
|---|---|---|---|---|---|
| **Average** | 35.46% | 31.49% | 27.96% | 29.11% | 40.49% |

**Table 4.** δ values for LCS

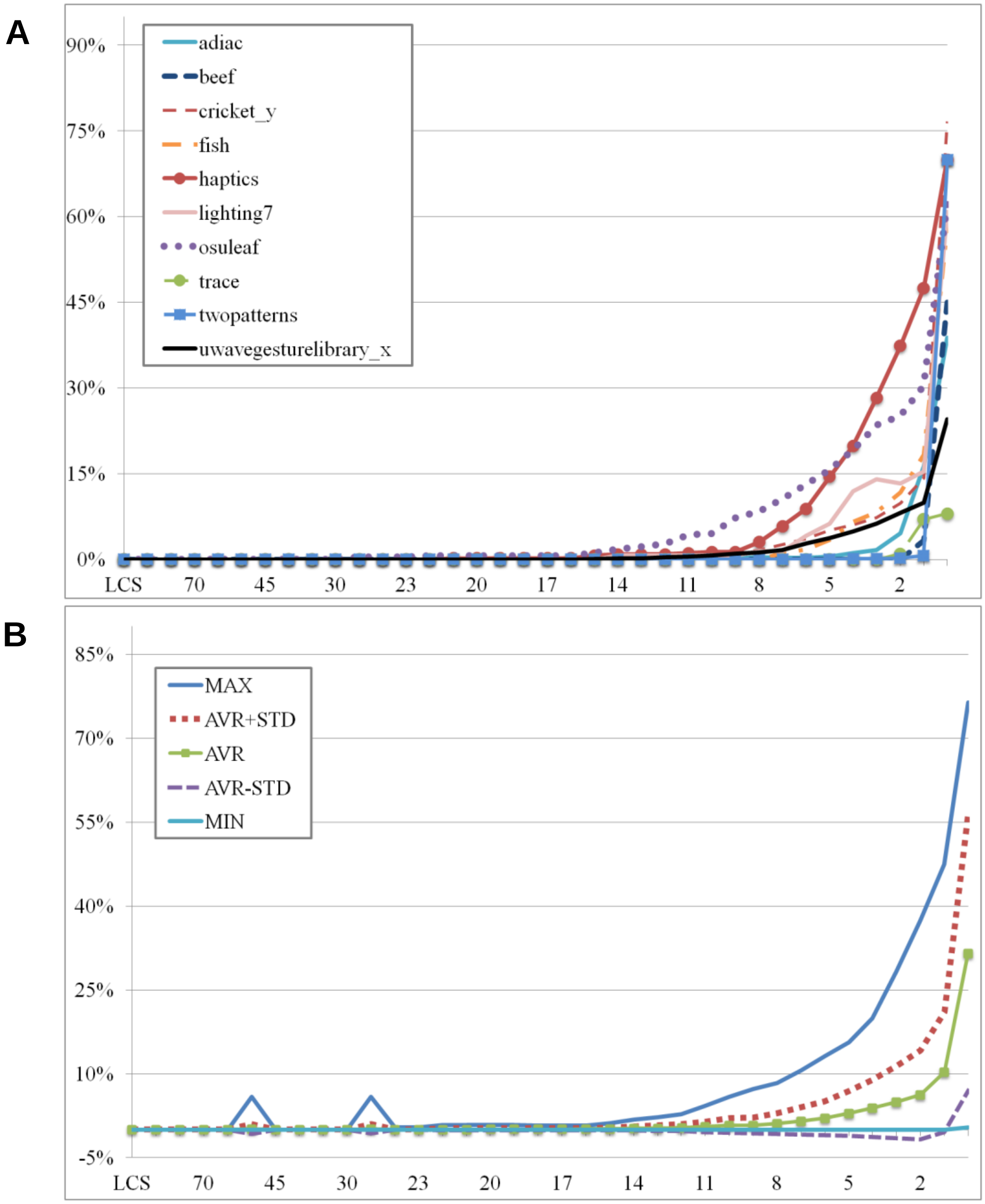

**Figure 12.** Change of classes for LCS

Changes in the case of ERP (Figure 13) start from around *r* = 60% and are most visible for *cricket*_y in the same manner as for the 1NN graph (Figure 6). Percentage of nodes in the nearest neighbor graph whose neighbors changed their classes compared to the unconstrained ERP reaches values higher than 50% for a smaller number of data sets (*cricket_x*, *cricket_y*, *cricket_z*, and *lighting7*). There are data sets (*cbf*, *cinc_ecg_torso*, *ecgfivedays*, *facefour*, *starlightcurves*, *twoleadecg*, *twopatterns*, *yoga*) for which we noticed significant changes in the nearest neighbor graph (higher than 60% for *r* = 0%) but that produce only minor changes of classes (less than 8%). The average value of $\delta$ decreases as the warping window becomes smaller. *Adiac*, *haptics* and *inlineskate* are still within the group of data sets which have the highest percentage of nodes with altered classes among the nodes changed by the Sakoe-Chiba band. We can see that several data sets with the lowest $\delta$ values reappear also with ERP.

| Data set | 4% | 3% | 2% | 1% | 0% | Data set | 4% | 3% | 2% | 1% | 0% |
|---|---|---|---|---|---|---|---|---|---|---|---|
| 50words | 52.96 | 48.21 | 44.95 | 42.12 | 43.88 | mallat | - | - | - | 0.00 | 2.51 |
| ● **adiac** | **-** | **100.00** | **100.00** | **80.00** | **59.54** | medicalimages | 37.93 | 37.90 | 33.67 | 35.94 | 35.94 |
| ● **beef** | **57.14** | **62.50** | **62.50** | **60.00** | **60.00** | motes | 13.33 | 13.25 | 12.37 | 12.96 | 12.96 |
| car | 100.00 | 33.33 | 22.22 | 27.59 | 29.17 | noninvasivefatalecg_thorax1 | 75.00 | 64.00 | 49.15 | 36.90 | 22.87 |
| ○ *cbf* | *0.00* | *0.00* | *0.00* | *0.00* | *2.60* | noninvasivefatalecg_thorax2 | 87.50 | 87.50 | 68.75 | 31.36 | 15.21 |
| chlorineconcentration | - | - | - | - | 0.00 | oliveoil | - | - | - | - | - |
| cinc_ecg_torso | 28.26 | 28.00 | 9.20 | 4.18 | 7.58 | osuleaf | 58.23 | 53.20 | 51.00 | 45.64 | 48.88 |
| coffee | - | - | - | - | - | plane | 0.00 | 50.00 | 8.33 | 5.88 | 4.50 |
| ● **cricket_x** | **72.92** | **69.85** | **68.64** | **66.80** | **74.34** | ○ *sonyaiborobotsurface* | *4.88* | *4.88* | *3.55* | *3.31* | *3.31* |
| ● **cricket_y** | **89.89** | **88.43** | **87.96** | **88.39** | **91.02** | ○ *sonyaiborobotsurfaceii* | *5.80* | *4.23* | *4.23* | *3.47* | *3.47* |
| ● **cricket_z** | **68.99** | **63.83** | **61.21** | **61.31** | **70.30** | starlightcurves | 13.14 | 14.01 | 13.77 | 12.76 | 10.56 |
| diatomsizereduction | 0.00 | 50.00 | 33.33 | 50.00 | 1.15 | swedishleaf | 16.28 | 13.37 | 13.56 | 20.60 | 25.33 |
| ecg200 | 59.26 | 45.95 | 29.85 | 17.65 | 17.65 | symbols | 10.89 | 8.57 | 5.63 | 5.01 | 5.43 |
| ○ *ecgfivedays* | *0.00* | *0.00* | *0.98* | *0.88* | *0.73* | ○ *synthetic_control* | *5.47* | *5.02* | *5.02* | *19.19* | *19.19* |
| faceall | 8.59 | 5.13 | 3.64 | 5.62 | 19.87 | trace | - | - | 0.00 | 8.70 | 14.75 |
| facefour | 50.00 | 20.00 | 5.88 | 4.17 | 4.23 | ○ *twoleadecg* | *1.33* | *1.37* | *0.30* | *0.93* | *0.93* |
| fish | 50.00 | 14.29 | 28.57 | 28.38 | 28.65 | ○ *twopatterns* | *0.00* | *0.00* | *0.16* | *0.49* | *2.72* |
| gun_point | - | - | 50.00 | 11.11 | 7.46 | uwavegesturelibrary_x | 33.42 | 30.96 | 29.41 | 28.93 | 29.25 |
| ● **haptics** | **63.64** | **66.94** | **63.64** | **62.71** | **62.85** | uwavegesturelibrary_y | 40.48 | 39.08 | 38.34 | 37.82 | 38.82 |
| ● **inlineskate** | **57.36** | **54.40** | **53.79** | **55.65** | **59.77** | uwavegesturelibrary_z | 36.51 | 36.36 | 34.44 | 33.93 | 34.35 |
| ○ *italypowerdemand* | *8.70* | *8.70* | *8.70* | *8.70* | *8.70* | wafer | 63.16 | 40.48 | 26.09 | 4.97 | 1.60 |
| lighting2 | 50.65 | 51.76 | 48.89 | 45.54 | 41.03 | wordssynonyms | 45.29 | 42.06 | 37.65 | 36.02 | 37.93 |
| ● **lighting7** | **90.63** | **90.00** | **88.35** | **90.16** | **91.91** | yoga | 22.73 | 22.12 | 16.81 | 11.03 | 8.13 |

| | 4% | 3% | 2% | 1% | 0% |
|---|---|---|---|---|---|
| **Average** | 37.96% | 36.74% | 31.54% | 28.07% | 26.39% |

**Table 5.** $\delta$ values for ERP

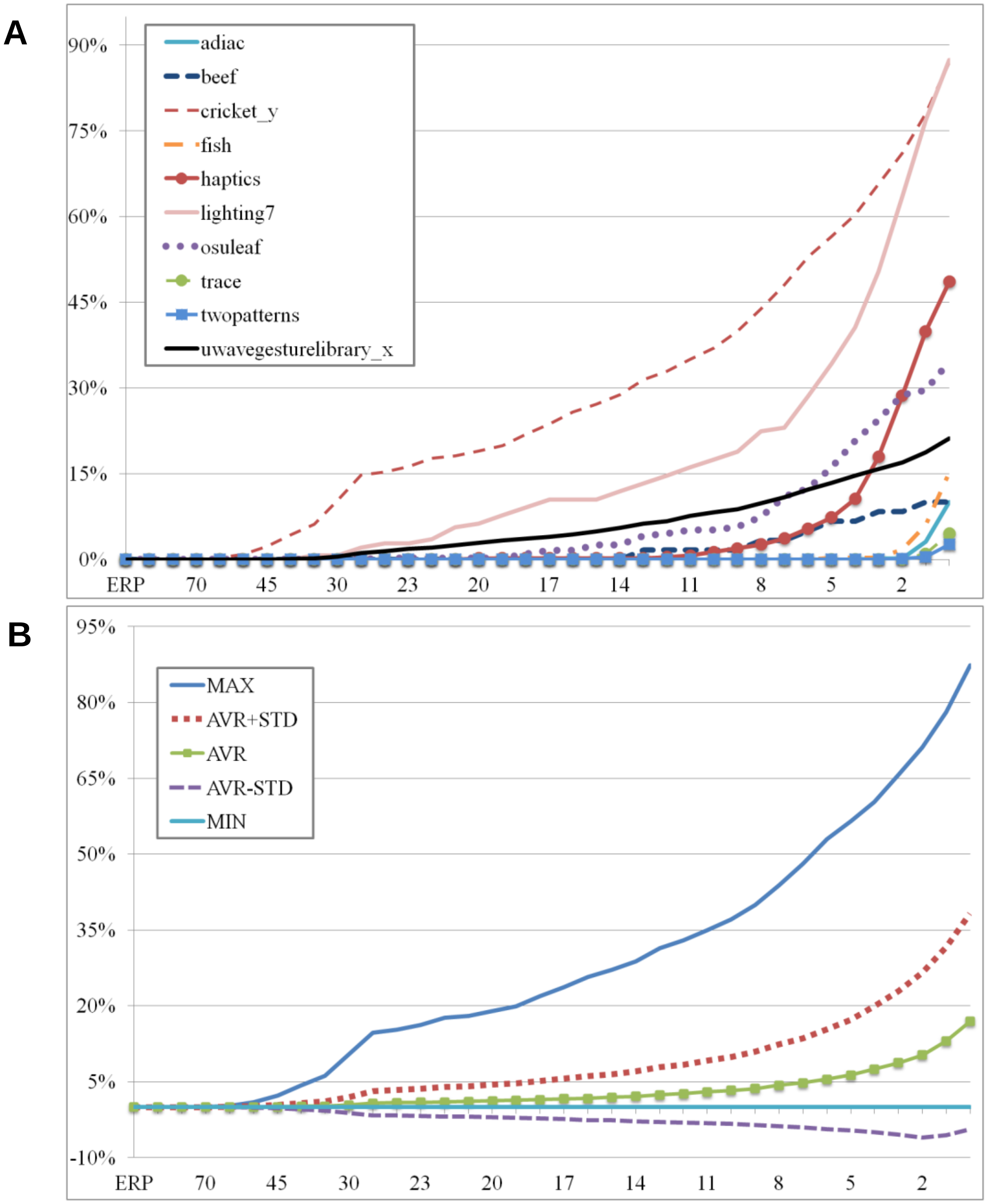

**Figure 13.** Change of classes for ERP

In compliance with the results for the EDR measure from Section 4.1, nodes with changed classes begin to appear when the width of the warping window is reduced below 20% of the length of the time series (Figure 14). Changes greater than one percent arise only for $r$ < 12%. For constraint values close to zero, the highest number of nodes with changed classes emerges in case of *haptics* and *inlineskate*. They are the only two data sets that achieve changes larger than 40%. In terms of results obtained for $\delta$, EDR most closely resembles ERP: they have very similar values and for both of them $\delta$ decreases as we reduce the width of the global constraint (the Wilcoxon sign-rank test reveals no significant difference). *Adiac*, *haptics* and *inlineskate* retain their place among the data sets with the highest δ values, and a larger number of data sets with very small changes among the classes are repeated for EDR too. Interestingly, only EDR generates large δ values for the *symbols* data set, all the other similarity measures yield values less than 11% (the only exception is LCS with *r* = 0%, which gives almost 40%).

| Data set | 4% | 3% | 2% | 1% | 0% | Data set | 4% | 3% | 2% | 1% | 0% |
|---|---|---|---|---|---|---|---|---|---|---|---|
| 50words | 40.95 | 37.99 | 34.26 | 31.91 | 34.00 | mallat | - | - | - | - | 0.00 |
| ● adiac | **75.00** | **62.50** | **85.71** | **66.67** | **53.26** | medicalimages | 29.41 | 32.54 | 37.43 | 42.83 | 42.83 |
| ● beef | **-** | **100.00** | **100.00** | **66.67** | **83.33** | ○ *motes* | *9.84* | *9.62* | *7.77* | *7.38* | *7.38* |
| car | 71.43 | 31.58 | 20.51 | 19.74 | 31.18 | noninvasivefatalecg_thorax1 | 80.00 | 85.71 | 76.92 | 65.75 | 28.11 |
| ○ *cbf* | *0.98* | *0.68* | *0.71* | *1.15* | *3.17* | noninvasivefatalecg_thorax2 | 50.00 | 66.67 | 80.00 | 54.76 | 17.63 |
| chlorineconcentration | - | - | - | - | 0.00 | oliveoil | - | - | - | - | 66.67 |
| ○ *cinc_ecg_torso* | - | - | *0.00* | *0.00* | *0.20* | osuleaf | 31.61 | 31.36 | 33.08 | 37.92 | 48.69 |
| coffee | - | 0.00 | 0.00 | 100.00 | 28.57 | ○ *plane* | *0.00* | *0.00* | *0.00* | *0.00* | *0.93* |
| cricket_x | 55.56 | 55.07 | 44.52 | 40.96 | 42.90 | sonyaiborobotsurface | 25.00 | 25.00 | 4.76 | 5.94 | 5.94 |
| cricket_y | 44.83 | 50.00 | 41.09 | 39.22 | 38.70 | ○ *sonyaiborobotsurfaceii* | *0.00* | *7.14* | *7.14* | *7.64* | *7.64* |
| cricket_z | 57.50 | 53.57 | 44.79 | 40.82 | 41.33 | starlightcurves | 12.80 | 14.57 | 14.33 | 13.17 | 12.33 |
| diatomsizereduction | 100.00 | 50.00 | 50.00 | 5.00 | 1.01 | swedishleaf | 8.75 | 9.55 | 13.25 | 17.25 | 26.04 |
| ecg200 | 33.33 | 37.50 | 25.00 | 10.87 | 10.87 | symbols | 66.67 | 42.86 | 42.86 | 32.00 | 28.33 |
| ecgfivedays | - | - | - | - | 0.00 | synthetic_control | 13.51 | 18.10 | 18.10 | 20.12 | 20.12 |
| ○ *faceall* | *7.84* | *5.97* | *3.60* | *4.24* | *5.02* | trace | - | - | 0.00 | 47.83 | 33.90 |
| facefour | 50.00 | 11.11 | 4.17 | 2.08 | 5.00 | ○ *twoleadecg* | - | *0.00* | *0.00* | *1.31* | *1.31* |
| fish | 47.62 | 28.21 | 27.84 | 24.88 | 35.07 | ○ *twopatterns* | *0.08* | *0.21* | *0.27* | *0.72* | *1.71* |
| ○ *gun_point* | *0.00* | *0.00* | *10.00* | *4.92* | *5.05* | uwavegesturelibrary_x | 46.82 | 40.94 | 34.79 | 30.56 | 29.22 |
| ● haptics | **70.79** | **66.43** | **61.99** | **65.13** | **65.14** | uwavegesturelibrary_y | 37.24 | 36.27 | 35.58 | 33.70 | 35.52 |
| ● inlineskate | **59.76** | **58.36** | **56.74** | **58.89** | **66.16** | uwavegesturelibrary_z | 33.84 | 32.56 | 30.84 | 32.08 | 35.34 |
| ○ *italypowerdemand* | *5.56* | *5.56* | *5.56* | *5.56* | *5.56* | wafer | 50.00 | 50.00 | 28.57 | 4.85 | 0.88 |
| lighting2 | - | 100.00 | 66.67 | 50.00 | 27.03 | wordssynonyms | 48.05 | 48.25 | 45.29 | 43.28 | 41.31 |
| lighting7 | - | 50.00 | 50.00 | 30.43 | 40.00 | yoga | 18.95 | 17.55 | 12.27 | 9.85 | 9.75 |

| | 4% | 3% | 2% | 1% | 0% |
|---|---|---|---|---|---|
| **Average** | 36.68% | 34.33% | 29.91% | 28.05% | 24.44% |

**Table 6.** δ values for EDR

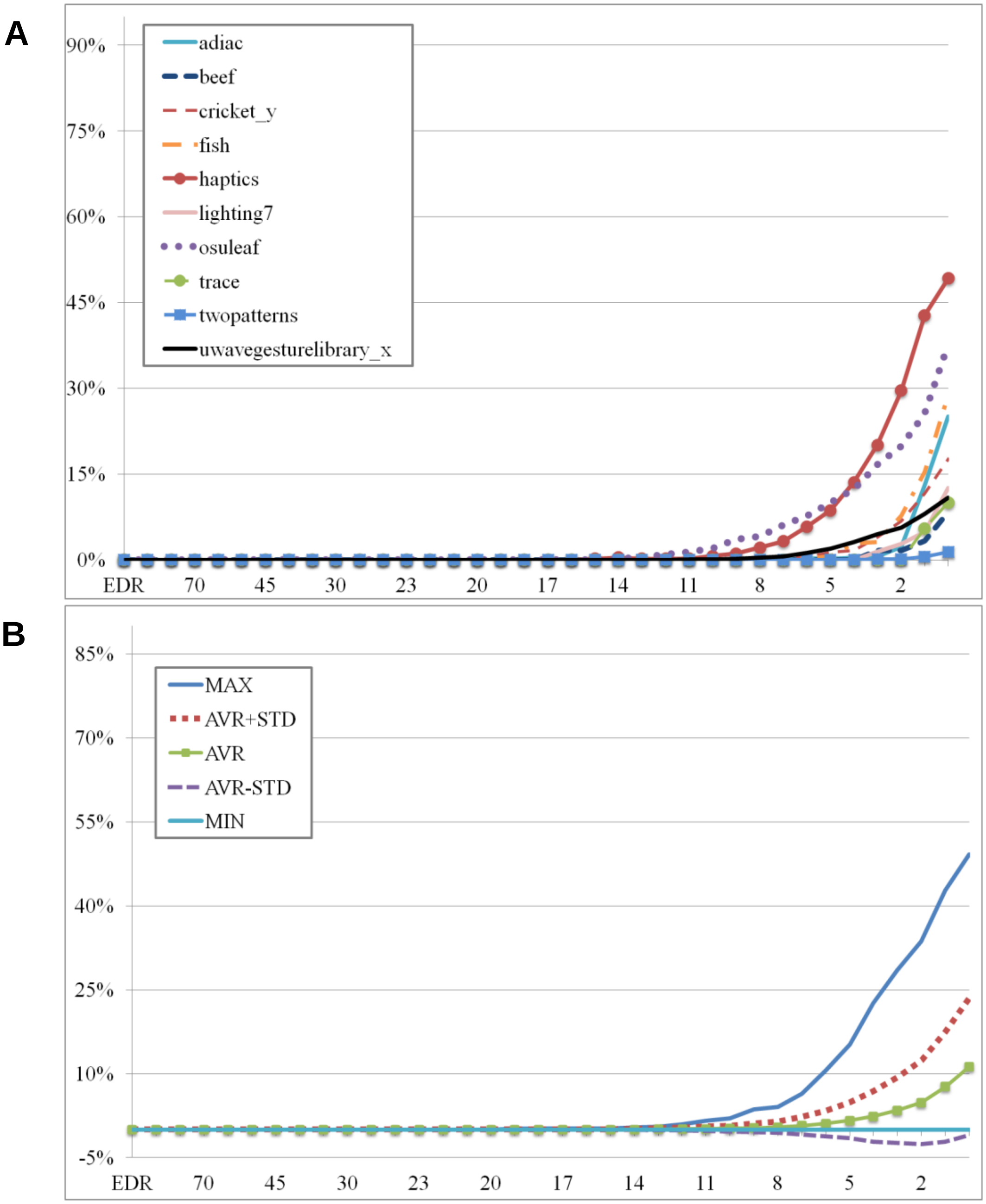

**Figure 14.** Change of classes for EDR

In the second part of the experiments we have seen that the properties identified by analyzing the structure of the nearest neighbor graphs are, in general, replicated among the data describing the change of classes. The average numbers of changed classes in 1NN graphs are presented in Figure 15. We can notice resemblance to the graphs in Figure 8 which shows the average number of changed nodes in the nearest neighbor graphs: the biggest changes are induced for DTW, the least ones for EDR, while LCS and ERP are in between. However, there are some differences, too. LCS and ERP provide similar average number of changed neighbors, but the average number of altered classes is higher for ERP. In terms of classes, ERP is closer to DTW and LCS is closer to EDR. The other significant difference is found for small values of *r* (< 2%). In this area the number of nodes for LCS that have changed their classes rapidly grows, and for *r* = 0% LCS overtakes DTW. This confirms that warping window of width *r* = 0% has a special impact on LCS.

Looking at the percentage of data sets for which there are nodes in the 1NN graph with changed classes, as shown in Figure 16, we can conclude that in case of all four measures the altered classes are beginning to appear immediately with the first changes in the structure of the nearest-neighbor graph. These graphs differ from the graphs shown in Figure 9 only slightly. This confirms that the Sakoe-Chiba band affects DTW the most, and that among the discussed similarity measures EDR is least affected.

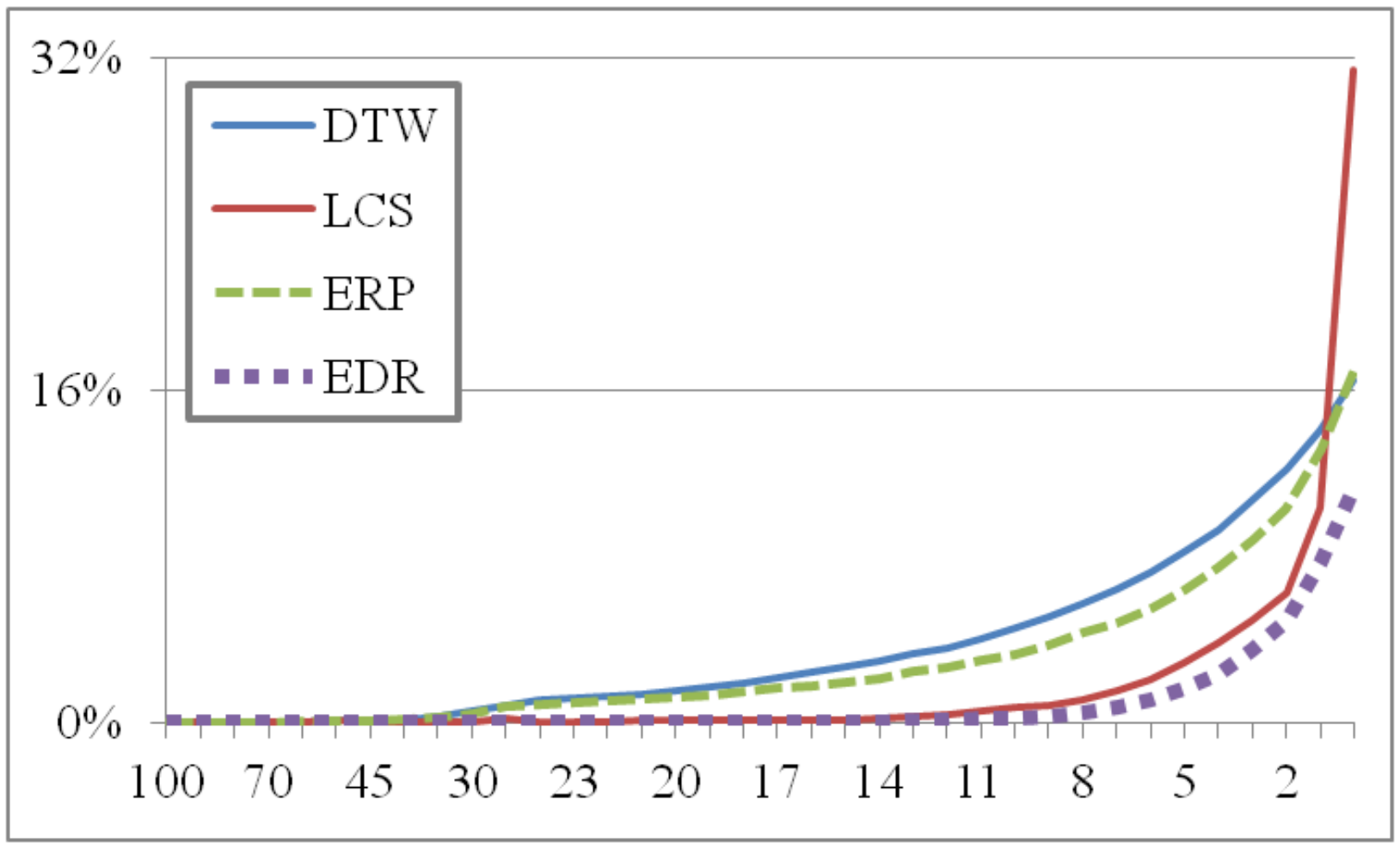

**Figure 15.** The average changes of classes

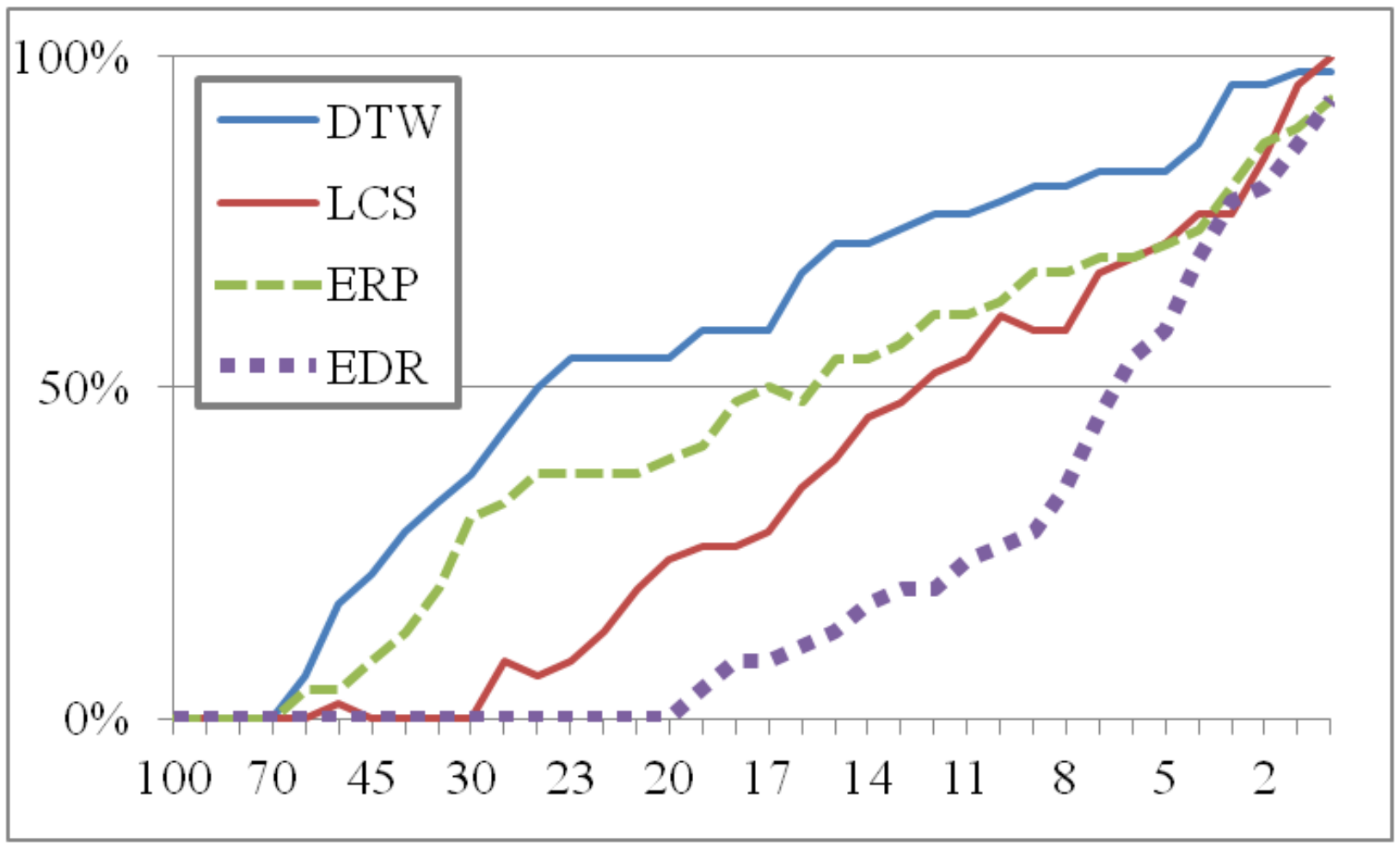

**Figure 16.** The percentage of the data sets with changed classes

By analyzing the percentage of those nodes in the 1NN graph which have altered their classes compared to the unconstrained measures, in relation to the total number of changed nodes (regardless of the class), we have seen that (for $r$ < 5%) there are groups of data sets whose members exhibit similar characteristics under the influence of the Sakoe-Chiba band independent of the similarity measure. The first such group consists of the *adiac*, *haptics* and *inlineskate* data sets. Global constraints change their 1NN graphs in such way that a large part (over 50%) of the new neighbors have altered classes. The second group includes *cbf*, *italypowerdemand*, *twoleadecg* and *twopatterns*. These data sets give the lowest $\delta$ values (up to 10%) for each of the four measures (the only exception occurs in *cbf* and *twopatterns* with LCS for $r$ = 0%).

In Section 4.1 we have seen that there is a group of data sets (including *chlorineconcentration*, *mallat*, *oliveoil* and *trace*) for which LCS, ERP and EDR show only slight changes in the structure of the 1NN graph (or no changes at all) but DTW demonstrates more significant ones. For most of these data sets DTW has very low delta values, this could be the explanation why it has smaller average $\delta$ values than the other measures despite the fact that DTW produces the biggest changes in the nearest neighbor graph and among the classes, too.

## 4.3 Impact on Classification Accuracy

The results of Section 4.1 and Section 4.2 clearly confirmed that the use of the Sakoe-Chiba global constraint causes recognizable changes in the behavior of the four popular elastic similarity measures, especially for small values of the warping window width. The last phase of our experiments is devoted to analyzing how these changes affect the accuracy of the 1NN classifier. In the first step we will use stratified 9-fold cross-validation (SCV1x9) to find the smallest values of parameter $r$ for which 1NN produces the smallest classification errors. In the second step, we will discuss to what extent the 1NN graphs that correspond to the values of parameter $r$ selected in the first step differ from the 1NN graphs of the unconstrained similarity measures. In the third step, we will give a comparative review of the observed similarity measures based on the classification accuracies obtained by 10 runs of stratified 10-fold cross-validation (SCV10x10) using the values of parameter $r$ from the first step. We will conclude our analysis in the fourth step by observing some general differences between the studied similarity measures.

The lowest widths of the warping window for which SCV1x9 gives the smallest classification error are presented in Table 7 (the same set of $r$ values was searched as in Section 4.1 and Section 4.2; in case of ties we report the lowest values of $r$). On average, this value is lowest in the case of DTW (about 4% of the length of time series) and highest for ERP (almost 10% of the length of time series). Values greater than 10% were observed only for some of the 46 investigated data sets. We have found only two such data sets for DTW (*lighting2* and *motes*), six for EDR, nine for ERP, and ten for LCS – whereas in case of ERP several data sets have values greater than 30% (*cinc_ecg_torso*, *cricket_x*, *cricket_y*, *cricket_z*, *lighting2*, *lighting7*, *motes*).

Overall, in most cases the results obtained for different similarity measures are close, but there are also data sets for which certain similarity measures differ substantially from the others in this respect (*cinc_ecg_torso*, *cricket_z*, *lighting2*, *lighting7*, *osuleaf*).

| Data set | DTW | LCS | ERP | EDR | Data set | DTW | LCS | ERP | EDR |
|---|---|---|---|---|---|---|---|---|---|
| 50words | 6 | 8 | 4 | **15** | mallat | 4 | 1 | 1 | 0 |
| adiac | 1 | 1 | 1 | 1 | medicalimages | 5 | 10 | 7 | 6 |
| beef | 2 | 2 | 10 | 4 | motes | **21** | **12** | **35** | **14** |
| car | 1 | 8 | 5 | 7 | noninvasivefatalecg_thorax1 | 0 | 1 | 2 | 6 |
| cbf | 2 | 8 | 1 | 5 | noninvasivefatalecg_thorax2 | 0 | 1 | 3 | 0 |
| chlorineconcentration | 0 | 0 | 0 | 0 | oliveoil | 0 | 2 | 0 | 1 |
| cinc_ecg_torso | 2 | 2 | **35** | 1 | osuleaf | 5 | **23** | **11** | **14** |
| coffee | 3 | 3 | 0 | 2 | plane | 5 | 5 | 4 | 1 |
| cricket_x | 7 | **12** | **40** | 9 | sonyaiborobotsurface | 0 | 3 | 5 | 5 |
| cricket_y | 10 | **14** | **70** | **12** | sonyaiborobotsurfaceii | 2 | 2 | 2 | 2 |
| cricket_z | 7 | **15** | **35** | 9 | starlightcurves | 10 | **14** | 2 | 6 |
| diatomsizereduction | 0 | 1 | 0 | 1 | swedishleaf | 3 | 5 | 3 | 8 |
| ecg200 | 0 | 2 | 0 | 4 | symbols | 7 | 6 | 5 | 6 |
| ecgfivedays | 1 | 1 | 0 | 0 | synthetic_control | 9 | **17** | 7 | 7 |
| faceall | 3 | 7 | 6 | 4 | trace | 4 | 4 | 2 | 2 |
| facefour | 3 | 1 | 9 | 4 | twoleadecg | 3 | 2 | **11** | 2 |
| fish | 1 | 5 | 3 | 6 | twopatterns | 4 | 5 | 3 | 5 |
| gun_point | 4 | 6 | 3 | 5 | uwavegesturelibrary_x | 6 | 9 | 7 | **12** |
| haptics | 5 | 6 | 4 | 6 | uwavegesturelibrary_y | 6 | 3 | 3 | 8 |
| inlineskate | 6 | 10 | 5 | 9 | uwavegesturelibrary_z | 3 | **16** | 4 | **15** |
| italypowerdemand | 0 | 5 | 0 | 0 | wafer | 0 | 4 | 3 | 0 |
| lighting2 | **13** | **15** | **45** | 1 | wordssynonyms | 6 | **12** | 5 | 7 |
| lighting7 | 5 | 5 | **40** | 4 | yoga | 2 | 7 | 5 | 7 |

| | DTW | LCS | ERP | EDR |
|---|---|---|---|---|
| **Average** | 4.07 | 6.54 | 9.70 | 5.28 |

**Table 7.** The values of parameter *r* for which SCV1x9 give the smallest error rate

In Section 4.1 and Section 4.2 we have shown that application of the Sakoe-Chiba band exerts the greatest influence on DTW and the lowest influence on EDR, while the magnitude of its effect on LCS and ERP is somewhere between these two boundary cases. This difference can be perceived among the data in Table 8, too. Table 8 contains the percentages of those nodes of 1NN graphs which have changed their classes under the influence of the constraint parameter *r* (whose values are taken from Table 7), compared to the nodes of the 1NN graphs of the unconstrained measures. The number of data sets for which SCV1x9 gives lowest classification error without changes in the NN graph (regarding the classes of the nodes) is smallest for DTW (4) and highest for EDR (25). For LCS and ERP there are 16 such data sets. In order to achieve the best classification accuracy, changes over 10% are most frequently required for DTW (20 data sets), followed by ERP (7 data sets). In case of LCS and EDR, changes of this magnitude are needed only for the *adiac* data set.

| Data set | DTW | LCS | ERP | EDR | Data set | DTW | LCS | ERP | EDR |
|---|---|---|---|---|---|---|---|---|---|
| 50words | *27.96* | 5.75 | *19.78* | **0.00** | mallat | 0.67 | **0.00** | **0.00** | **0.00** |
| adiac | *14.47* | *16.39* | 3.07 | *13.06* | medicalimages | *10.60* | 2.10 | 3.59 | 1.49 |
| beef | 3.33 | **0.00** | 1.67 | **0.00** | motes | 1.89 | 0.08 | **0.00** | **0.00** |
| car | *18.33* | **0.00** | **0.00** | **0.00** | noninvasivefatalecg_thorax1 | *21.70* | 1.35 | 0.77 | 0.03 |
| cbf | **0.00** | **0.00** | **0.00** | **0.00** | noninvasivefatalecg_thorax2 | *13.25* | 1.33 | 0.19 | 7.92 |
| chlorineconcentration | **0.00** | 0.37 | **0.00** | **0.00** | oliveoil | 3.33 | **0.00** | **0.00** | **0.00** |
| cinc_ecg_torso | 1.41 | **0.00** | **0.00** | **0.00** | osuleaf | *24.21* | 0.45 | 5.20 | 0.23 |
| coffee | 5.36 | **0.00** | **0.00** | **0.00** | plane | **0.00** | **0.00** | **0.00** | **0.00** |
| cricket_x | *13.21* | 0.51 | 0.38 | **0.00** | sonyaiborobotsurface | 1.93 | **0.00** | 0.81 | **0.00** |
| cricket_y | *10.26* | 0.13 | **0.00** | **0.00** | sonyaiborobotsurfaceii | 2.76 | 1.63 | 1.43 | 0.10 |
| cricket_z | *13.08* | 0.13 | 0.26 | **0.00** | starlightcurves | 1.93 | 0.17 | 2.45 | 0.10 |
| diatomsizereduction | 0.31 | 0.62 | 0.31 | 0.31 | swedishleaf | *15.29* | 0.98 | 2.22 | **0.00** |
| ecg200 | *12.50* | 7.50 | *10.50* | 1.50 | symbols | 0.29 | 0.78 | 0.39 | **0.00** |
| ecgfivedays | 0.68 | **0.00** | 0.57 | **0.00** | synthetic_control | 0.83 | **0.00** | 2.17 | **0.00** |
| faceall | 1.73 | 0.04 | 0.49 | 0.18 | trace | **0.00** | **0.00** | **0.00** | **0.00** |
| facefour | 2.68 | 1.79 | 0.89 | 0.89 | twoleadecg | 0.09 | **0.00** | **0.00** | **0.00** |
| fish | *21.14* | 3.43 | 0.29 | 0.29 | twopatterns | **0.00** | **0.00** | **0.00** | **0.00** |
| gun_point | 8.00 | **0.00** | **0.00** | **0.00** | uwavegesturelibrary_x | *20.12* | 0.96 | *10.92* | **0.00** |
| haptics | *30.89* | 8.86 | *10.58* | 5.83 | uwavegesturelibrary_y | *26.66* | 3.04 | *16.66* | 1.25 |
| inlineskate | *22.46* | 0.46 | 8.77 | 2.62 | uwavegesturelibrary_z | *26.51* | 0.76 | *14.96* | 0.18 |
| italypowerdemand | 2.65 | 0.18 | 1.09 | 0.09 | wafer | 0.57 | 0.06 | 0.24 | 0.11 |
| lighting2 | 4.13 | **0.00** | **0.00** | 7.44 | wordssynonyms | *26.30* | 1.33 | *14.48* | 0.55 |
| lighting7 | *17.48* | 6.29 | **0.00** | **0.00** | yoga | 5.06 | 0.39 | 0.21 | 0.03 |

| | DTW | LCS | ERP | EDR |
|---|---|---|---|---|
| **Average** | 9.48% | 1.48% | 2.94% | 0.96% |

**Table 8.** Percentage of nodes in 1NN graph with changed classes for the values of parameter *r* from Table 7, compared to unconstrained measures

In order to compare the classification performance of the studied similarity measures we computed 1NN classification errors with the SCV10x10 evaluation method using the results from Table 7 as values for the global constraint parameter *r*. The lowest average error is produced by LCS (11.43%), the largest one by ERP (12.44%), and DTW (11.52%) and EDR (11.86%) are in between (Table 9). Looking at individual data sets the lowest classification error most often occurs with DTW (21 data sets), then with EDR (12 data sets) followed by LCS (11 data sets) and ERP (7 data sets). The mean value of the differences between the minimum and maximum errors is about 3.82. The biggest differences are with the following data sets: *symbols* (18.65), *osuleaf* (13.55), *car* (10.83), *lighting2* (8.79) and *trace* (8.35).

| Data set | DTW | LCS | ERP | EDR | Data set | DTW | LCS | ERP | EDR |
|---|---|---|---|---|---|---|---|---|---|
| 50words | 18.63 | 16.03 ● | 21.27 ○ | **15.51** ● | mallat | 1.20 | 7.02 ○ | **0.63** ● | 7.03 ○ |
| adiac | **31.42** | 33.18 ○ | 32.24 | 32.24 | medicalimages | **18.71** | 22.76 ○ | 18.72 | 21.81 ○ |
| beef | 47.67 | **43.00** | 51.17 ○ | 43.33 | motes | 4.51 | **1.45** ● | 2.95 ● | 1.71 ● |
| car | 18.17 | 11.42 ● | 20.17 | **9.33** ● | noninvasivefatalecg_thorax1 | **15.65** | 18.27 ○ | 16.49 ○ | 18.14 ○ |
| cbf | 0.02 | 0.02 | **0.00** | 0.05 | noninvasivefatalecg_thorax2 | **9.48** | 11.37 ○ | 9.61 | 10.87 ○ |
| chlorineconcentration | **0.27** | 0.81 ○ | 0.60 ○ | 0.80 ○ | oliveoil | **11.17** | 12.00 | 12.00 | 13.00 |
| cinc_ecg_torso | **0.01** | **0.01** | 0.23 ○ | 0.07 | osuleaf | 26.49 | **12.94** ● | 24.89 | 12.97 ● |
| coffee | 4.83 | **4.47** | 11.93 ○ | 4.57 | plane | **0.00** | 0.05 | 0.10 | 0.05 |
| cricket_x | **15.68** | 18.91 ○ | 20.29 ○ | 18.92 ○ | sonyaiborobotsurface | 1.29 | 1.55 | **0.90** | 1.66 |
| cricket_y | **13.73** | 15.72 ○ | 17.36 ○ | 16.88 ○ | sonyaiborobotsurfaceii | 1.19 | 1.82 ○ | **1.10** | **1.10** |
| cricket_z | **14.95** | 18.01 ○ | 20.55 ○ | 19.38 ○ | starlightcurves | **6.28** | 9.23 ○ | 10.03 ○ | 9.06 ○ |
| diatomsizereduction | **0.06** | **0.06** | **0.06** | **0.06** | swedishleaf | 11.32 | 8.54 ● | 9.88 ● | **7.99** ● |
| ecg200 | 10.00 | 9.10 | **7.35** ● | 9.40 | symbols | 1.62 | **1.36** | 1.71 | 20.01 ○ |
| ecgfivedays | 0.17 | **0.10** | 0.25 | 0.15 | synthetic_control | **0.45** | 2.18 ○ | 1.02 ○ | 3.02 ○ |
| faceall | 1.45 | 0.87 ● | 0.90 ● | **0.71** ● | trace | **0.10** | 0.25 | 8.45 ○ | 1.80 ○ |
| facefour | 3.66 | 1.08 ● | 0.98 ● | **0.89** ● | twoleadecg | **0.03** | 0.12 | 0.19 ○ | 0.11 |
| fish | 13.46 | 9.97 ● | 12.77 | **7.77** ● | twopatterns | **0.00** | 0.01 | 0.01 | 0.01 |
| gun_point | 1.90 | **0.60** ● | 1.75 | **0.60** ● | uwavegesturelibrary_x | **19.16** | 21.40 ○ | 20.68 ○ | 21.34 ○ |
| haptics | 53.50 | 51.34 | 54.14 | **51.31** | uwavegesturelibrary_y | **25.66** | 30.14 ○ | 27.97 ○ | 26.36 ○ |
| inlineskate | 44.18 | 40.80 ● | **36.45** ● | 40.31 ● | uwavegesturelibrary_z | 25.46 | **24.72** ● | 26.97 ○ | 26.63 ○ |
| italypowerdemand | 3.41 | 3.50 | 3.70 | **2.79** ● | wafer | **0.11** | 0.13 | 0.19 ○ | 0.12 |
| lighting2 | **9.50** | 14.54 ○ | 12.57 | 18.29 ○ | wordssynonyms | 17.43 | **15.51** ● | 19.05 ○ | 18.51 |
| lighting7 | **21.17** | 26.68 ○ | 28.37 ○ | 26.22 ○ | yoga | 4.66 | 2.88 ● | 3.60 ● | **2.58** ● |

| | DTW | LCS | ERP | EDR |
|---|---|---|---|---|
| Average | 11.52% | 11.43% | 12.44% | 11.86% |

**Table 9.** Classification errors obtained for SCV10x10
with the values of parameter *r* from Table 7

Statistically significant differences in error rates are denoted by symbols ● and ○ in Table 9, with the former signifying improvement, and the latter degradation of classifier performance when comparing LCS, ERP and EDR measures with DTW. For this we employed the corrected resampled t-test (Nadeau and Bengio, 2003) which adjusts to the loss in degrees of freedom due to repeated runs of cross-validation, at significance level 0.001. We report DTW as the baseline method in Table 9 since it is the recommended best choice of distance measure (Ding et al., 2008).

In order to assess whether some distance measure can be said to be better than others in the average case, we counted the statistically significant wins and losses according to the corrected resampled t-test, for each distance measure, with the results summarized in Table 10. The counts suggest that DTW is generally better (despite having slightly higher average error rate than LCS), with LCS and EDR tied second, and ERP exhibiting the worst performance.

| | Wins | Losses | W–L |
|---|---|---|---|
| **DTW** | 52 | 33 | 19 |
| **LCS** | 41 | 37 | 4 |
| **ERP** | 31 | 58 | –27 |
| **EDR** | 40 | 36 | 4 |

**Table 10.** Statistically significant wins and losses counts for the 1NN classifier with different distance measures, across all data sets

On the other hand, when we compare the average error rates across all data sets using the Wilcoxon sign-rank test (as in previous sections), the differences are not particularly strong, as shown in Table 11 which contains the corresponding *p* values. The one possibly significant difference when using this test is between DTW and ERP.

| | LCS | ERP | EDR |
|---|---|---|---|
| **DTW** | 0.79738 | 0.018043 | 0.4597 |
| **LCS** | | 0.09515 | 0.93264 |
| **ERP** | | | 0.12063 |

**Table 11.** *p* values for the pairwise Wilcoxon sign-rank test of the differences in average error rates across the data sets

Overall, based on the statistical tests, we can conclude that there is some evidence to consider DTW as the generally best distance measure, and ERP as the generally worst, but the evidence is not overwhelming. Furthermore, when observing the statistical differences on individual data sets (corrected resampled t-test, 0.001 significance level), for every distance measure there are at least a couple of data sets where the measure is significantly superior to all others. Therefore, the choice of the best distance measure for a particular problem may be different for the generally best case.

Observing the graphs of average classification errors across different values of *r* (Figure 17), the most evident common characteristic of the four discussed similarity measure is that for small widths of the warping window (< 6%) the average classification error steeply increases, and in all four cases reaches its maximum for $r = 0\%$ (DTW: 15.97%, LCS: 33.56%, ERP: 20.67%, EDR: 17.20%). The largest increase occurs for LCS and the lowest one for DTW. Within the area from $r = 100\%$ to $r = 6\%$ the similarity measures exhibit different behaviors. While in case of DTW the average classification error almost monotonously decreases from 14.04% to 12.38%, for ERP it almost monotonously increases from 13.03% to 14.63%. While a tendency of growth can also be noticed for LCS and EDR, the changes are very subtle: in case of LCS the average error ranges between 11.62% and 11.98%, and in case of EDR it ranges between 11.87% and 12.10%. This suggests that although DTW can be considered the best general choice according the previous analysis, LCS and EDR could be safer choices because of the less pronounced need for tuning the *r* parameter.

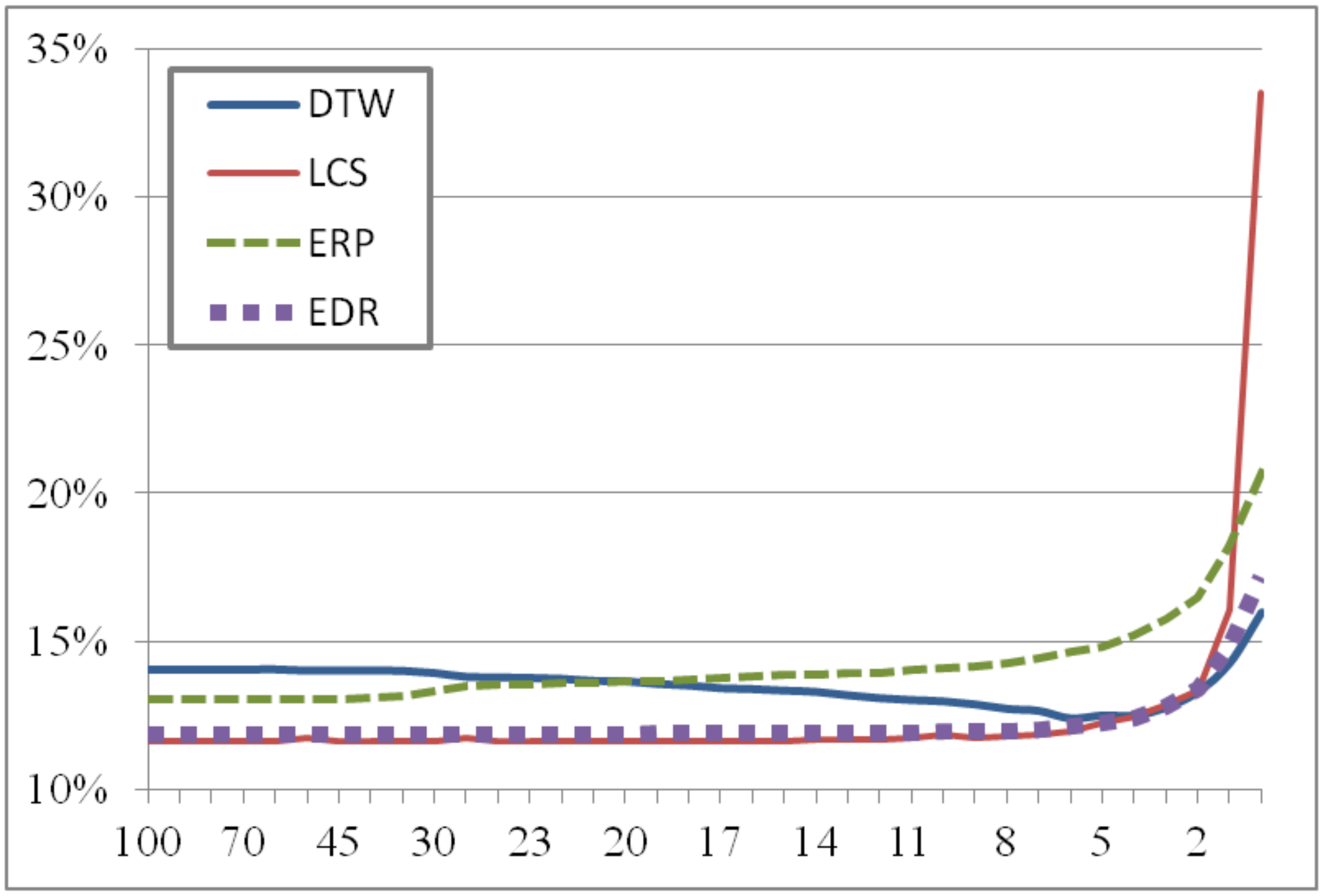

**Figure 17.** Average classification errors for SCV10x10

## 5. Conclusions and Future Work

A suitable choice of similarity measure between time series is an important part of similarity-based retrieval, and is in some form included in many tasks of time-series analysis. Since Euclidean distance is a very simple measure which is calculated quickly and represents a distance metric, it has become one of the most commonly used measures of similarity between time series (Agrawal et al., 1993; Chan and Fu, 1999; Keogh et al., 2001a, 2001b). However, it has two major disadvantages: it can only work with time series of the same length and is sensitive to distortions and shifting along the time axis (Keogh, 2002; Ratanamahatana and Keogh, 2005). To overcome these weaknesses many elastic measures are proposed in the literature (DTW, LCS, ERP, EDR, etc.). These measures have better classification accuracy than Euclidean distance (Ding et al., 2008), but they are all based on dynamic programming, which means that their computation complexity is quadratic. To decrease the computation time of the elastic measures global constraints are introduced, narrowing the search path in the matrix.

It was suggested that, in the case of DTW, the use of global constraints can actually improve the accuracy of classification compared to unconstrained similarity measures (Ratanamahatana and Keogh, 2005; Xi et al., 2006). In our previous work (Kurbalija et al., 2011), based on a smaller number of different warping window widths, we have determined that the constrained versions of the DTW and LCS measures qualitatively differ from their unconstrained counterparts and analyzed the speed-up gained by using global constraints. In this paper we have expanded our study of the impact of global constraints on the four most widely used elastic similarity measures: DTW, LCS, ERP and EDR. Through an extensive set of experiments we have described in detail the impact of the Sakoe-Chiba band on the nearest-neighbor graph. We showed that the constrained measures are qualitatively different than the unconstrained ones. From the obtained results we can clearly see that for low values of the constraint (less than 15%–10%) the change of the 1NN graph becomes significant for all of the considered similarity measures. In addition to this, the results reveal that there are notable differences in the effects of the constraints on different distance measures. DTW was found to be the most sensitive to the introduction of global constraints regarding the 1NN graph, while EDR is the least sensitive. The behavior of ERP and LCS measures was determined to be somewhere in between. Furthermore, comparison of 1NN classifier performance showed that DTW generally has a slight edge over other distance measures (especially ERP), but is more sensitive to the choice of the $r$ parameter.

The findings of our studies have clearly shown that all of the main elastic similarity measures (DTW, LCS, ERP and EDR) significantly change their behavior for small values of the global constraint. Thus, we expect our results to aid researchers and practitioners in selecting and tuning appropriate time-series similarity measures for their respective tasks, making the selection/tuning process simpler and faster, at the same time producing more accurate results. In addition, the insight into the behavior of similarity measures with respect to changing constraints can be beneficial to the design of efficient indexing strategies for fast computation of (approximate) nearest neighbors. In future work, we plan to expand the investigation of the effects of these changes on the accuracy of a wider range of distance-based classifiers. It would also be interesting to explore the influence of the Itakura parallelogram compared to the influence of the Sakoe-Chiba band.

## Acknowledgments

V. Kurbalija, M. Radovanović and M. Ivanović thank the Serbian Ministry of Education, Science and Technological Development for support through project no. OI174023, “Intelligent Techniques and their Integration into Wide-Spectrum Decision Support.”